\documentclass[letterpaper, 10 pt, journal, twoside]{ieeetran}

\usepackage{array,multirow,graphics}
\usepackage{graphicx}
\usepackage{times}
\usepackage{amsmath}
\usepackage{amssymb}
\usepackage{amsbsy}
\usepackage[font=footnotesize]{caption}
\usepackage{float}
\usepackage{balance}
\usepackage{url}
\usepackage[ruled,linesnumbered,longend]{algorithm2e}
\usepackage{xcolor}
\usepackage{latexsym}
\usepackage{caption}
\usepackage{subcaption}
\usepackage{textcomp}
\usepackage{pdfrender}
\usepackage{tabularx}
\usepackage{booktabs}
\usepackage{float}
\usepackage{gensymb}
\usepackage{mathtools}
\usepackage{orcidlink}

\usepackage{pdfpages} 
\usepackage{pgffor} 
\makeatletter
\AtBeginDocument{\let\LS@rot\@undefined}
\makeatother

\def\supplementfilename{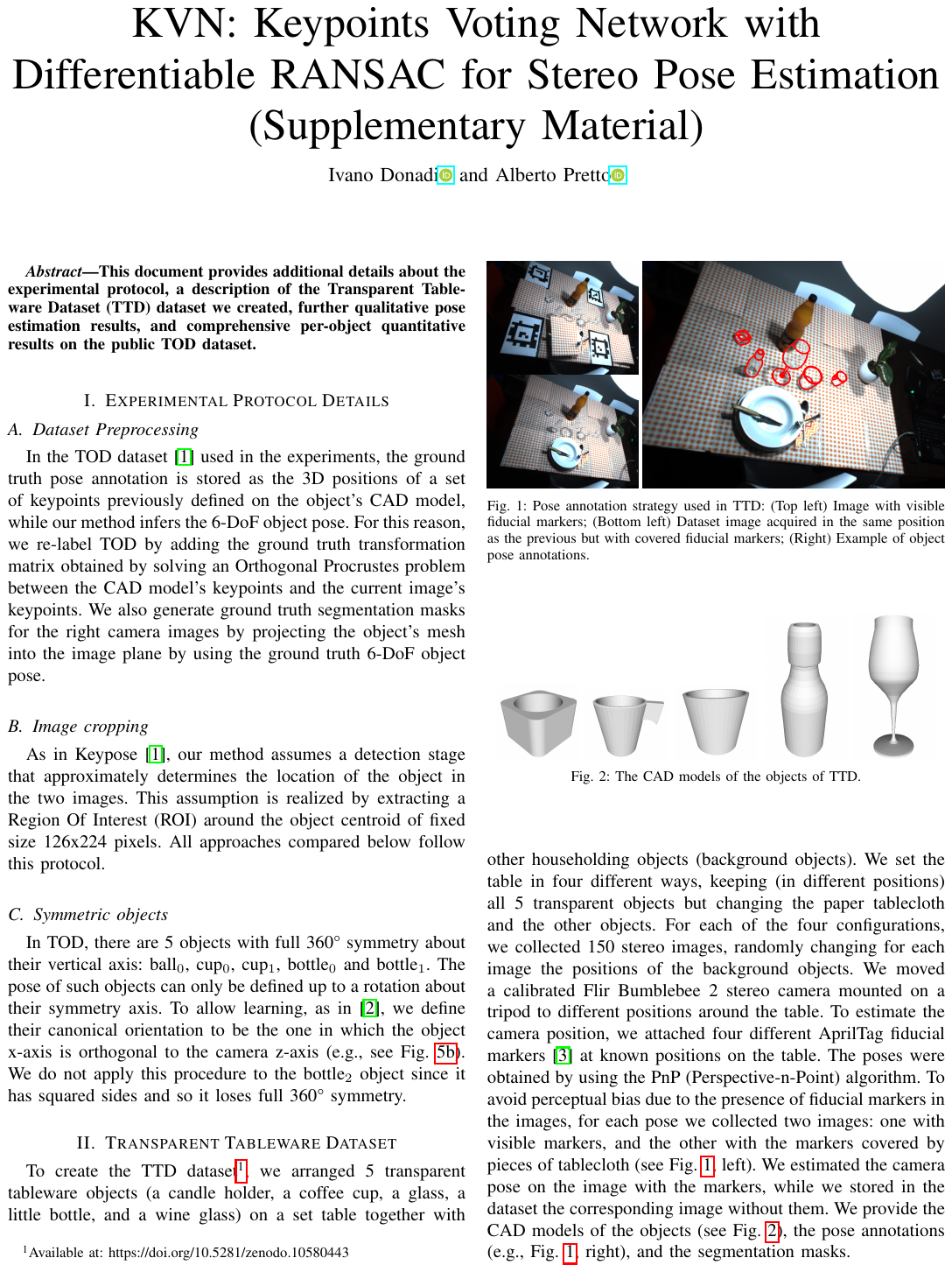}

\newif\ifarXiv
\arXivtrue 

\DeclareMathOperator*{\argmin}{arg\,min}

\IEEEoverridecommandlockouts                              

\usepackage{fancyhdr}
\fancypagestyle{arxivhdr}
{
   \fancyhf{}
   \setlength{\headheight}{15pt} 
\fancyfoot[C]{This paper has been accepted for publication in IEEE Robotics and Automation Letters\\DOI: 10.1109/LRA.2024.3367508 \url{https://ieeexplore.ieee.org/document/10440415}}
\fancyhead[C]{\footnotesize Please cite this paper as:\\
I. Donadi and A. Pretto, "KVN: Keypoints Voting Network with Differentiable RANSAC for Stereo Pose Estimation,"\\in IEEE Robotics and Automation Letters, vol. 9, no. 4, pp. 3498-3505, April 2024, doi: 10.1109/LRA.2024.3367508.}
}

\def\secref#1{Sec.~\ref{#1}}
\def\figref#1{Fig.~\ref{#1}}
\def\tabref#1{Tab.~\ref{#1}}
\def\eqref#1{Eq.~(\ref{#1})}

\title{KVN: Keypoints Voting Network with Differentiable RANSAC for Stereo Pose Estimation}

\author{Ivano Donadi\orcidlink{0009-0002-2362-066X} and Alberto Pretto\orcidlink{0000-0003-1920-2887}
\thanks{Manuscript received: July 16, 2023; Revised:
September 28, 2023; Accepted: January 8, 2024.}
\thanks{This paper was recommended for publication by
Editor Markus Vincze upon evaluation of the Associate Editor and Reviewers' comments.}
\thanks{This work was supported by the University of Padova under Grant UNI-IMPRESA-2020-SubEye.}
\thanks{The authors are with the Department of Information Engineering, University of Padova, Italy. Email: { \{ivano.donadi, alberto.pretto\}@dei.unipd.it}}
\thanks{Digital Object Identifier (DOI): see top of this page.}
}
\begin{document}

\maketitle
\thispagestyle{arxivhdr}


\begin{abstract}
Object pose estimation is a fundamental computer vision task exploited in several robotics and augmented reality applications. Many established approaches rely on predicting 2D-3D keypoint correspondences using RANSAC (Random sample consensus) and estimating the object pose using the PnP (Perspective-n-Point) algorithm. Being RANSAC non-differentiable,  correspondences cannot be directly learned in an end-to-end fashion. In this paper, we address the stereo image-based object pose estimation problem by i) introducing a differentiable RANSAC layer into a well-known monocular pose estimation network; ii) exploiting an uncertainty-driven multi-view PnP solver which can fuse information from multiple views. We evaluate our approach on a challenging public stereo object pose estimation dataset and a custom-built dataset we call Transparent Tableware Dataset (TTD), yielding state-of-the-art results against other recent approaches. Furthermore, in our ablation study, we show that the differentiable RANSAC layer plays a significant role in the accuracy of the proposed method. We release with this paper the code of our method and the TTD dataset.
\end{abstract}
\vspace{-3mm}
\section{Introduction}\label{sec:introduction}
\IEEEPARstart{A}{reliable} and accurate object pose estimation system is an essential requirement for many robotics applications, from robot-aided manufacturing to service robotics applications, as well as in several augmented reality tasks. Its goal is to estimate the 3D rotation and 3D translation of one or more known objects with respect to a canonical frame. 
This non-trivial task has been addressed extensively in recent decades but still raises open challenges that should be tackled \cite{thalhammer2023open}, like occlusion handling and localizing objects made of challenging materials (e.g., transparent materials).
The best-performing pose estimation methods rely on deep networks estimating the 6D object pose from single RGB images \cite{yolopose,9309178,Yu20206DoFOP,Song2020HybridPose6O,JIANG202216,9156435,Oberweger2018}, RGB-D images \cite{8953386,9093272,Wu2022} or stereo images \cite{liu2020keypose,9636459,chen2022stereopose}. Pose estimation from a single RGB image suffers from the lack of depth information, especially in challenging conditions like occlusions and cluttering. The depth map provided by RGB-D sensors represents a valid aid to reduce the search space and increase the robustness and the accuracy of the estimate. On the other hand, these methods are limited to opaque Lambertian objects which enable to obtain consistent depth information, while in the presence of reflective or transparent objects, the quality of the depth information is usually inadequate \cite{liu2020keypose,9197518}. In these cases, a calibrated RGB stereo rig is often a more suitable choice. 
\begin{figure}[t]
\centering
\begin{subfigure}{.5\linewidth}
  \centering
  \includegraphics[width=0.99\linewidth,trim={3cm 3cm 3cm 3cm},clip]{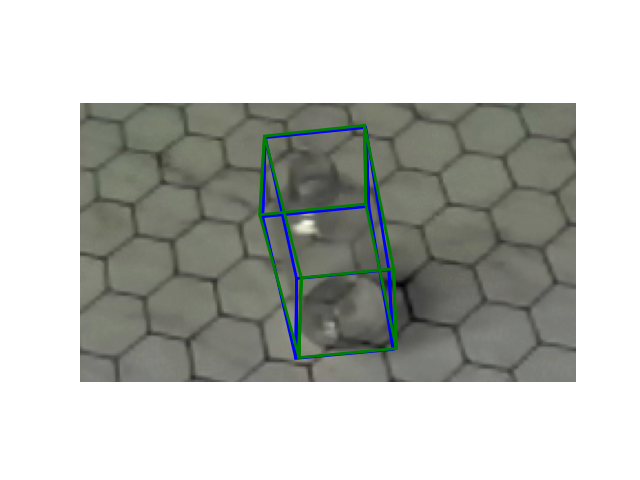}
  \caption{}
  \label{fig:teaser:bottle}
\end{subfigure}%
\begin{subfigure}{.5\linewidth}
  \centering
  \includegraphics[width=0.99\linewidth,trim={3cm 3cm 3cm 3cm},clip]{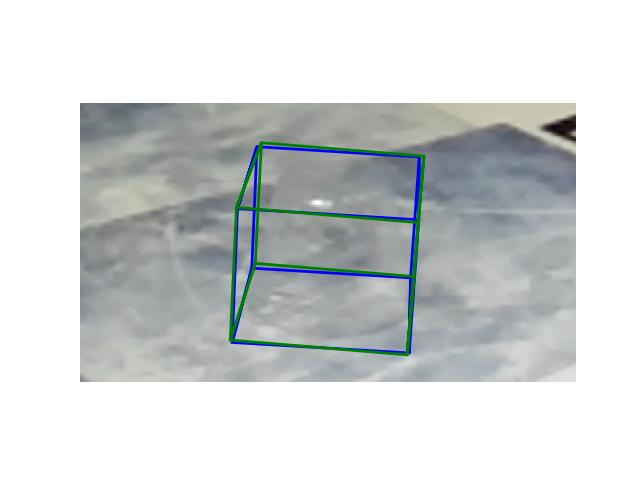}
  \caption{}
  \label{fig:teaser:cup}
\end{subfigure}\\
\begin{subfigure}{.5\linewidth}
  \centering
  \includegraphics[width=0.99\linewidth,trim={3cm 3cm 3cm 3cm},clip]{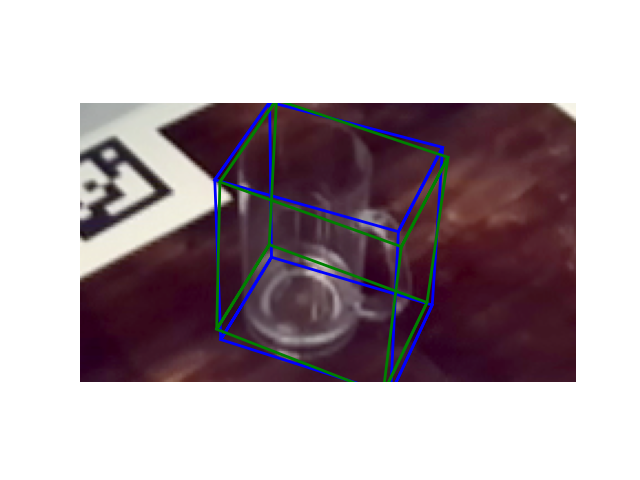}
  \caption{}
  \label{fig:teaser:mug}
\end{subfigure}%
\begin{subfigure}{.5\linewidth}
  \centering
  \includegraphics[width=0.99\linewidth,trim={3cm 3cm 3cm 3cm},clip]{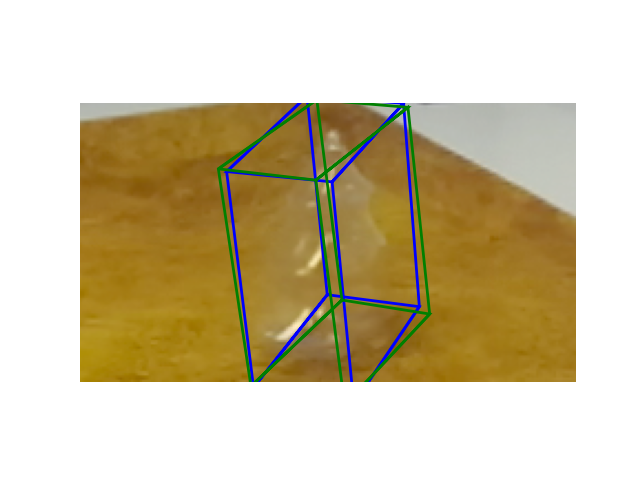}
  \caption{}
  \label{fig:teaser:tree}
\end{subfigure}\\
\begin{subfigure}{.5\linewidth}
  \centering
  \includegraphics[width=0.99\linewidth, height=3cm]{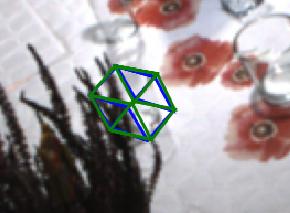}
  \caption{}
  \label{fig:ttd_result1}
\end{subfigure}%
\begin{subfigure}{.5\linewidth}
  \centering
  \includegraphics[width=0.99\linewidth]{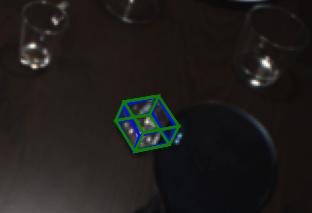}
  \caption{}
  \label{fig:ttd_result2}
\end{subfigure}
\caption{Predicted (blue) vs. ground truth (green) poses for KVN on 4 objects from the TOD dataset (a-d) and 2 objects from the TTD dataset (e,f). (\subref{fig:teaser:bottle}) and (\subref{fig:teaser:cup}) contain symmetric objects, respectively a small bottle and an upside-down cup, while (\subref{fig:teaser:mug}), (\subref{fig:teaser:tree}), (\subref{fig:ttd_result1}), and (\subref{fig:ttd_result2}) contain a mug, a Christmas tree, a coffee cup, and a candle holder, respectively. KVN can provide correct estimations even in cases where the object is barely distinguishable from the background, as in (\subref{fig:teaser:cup}), or occluded, as in (e,f).} 
\label{fig:teaser}
\vspace{-0.5cm}
\end{figure}
\begin{figure*}[t!]
   \centering
   \includegraphics[width=0.98\linewidth]{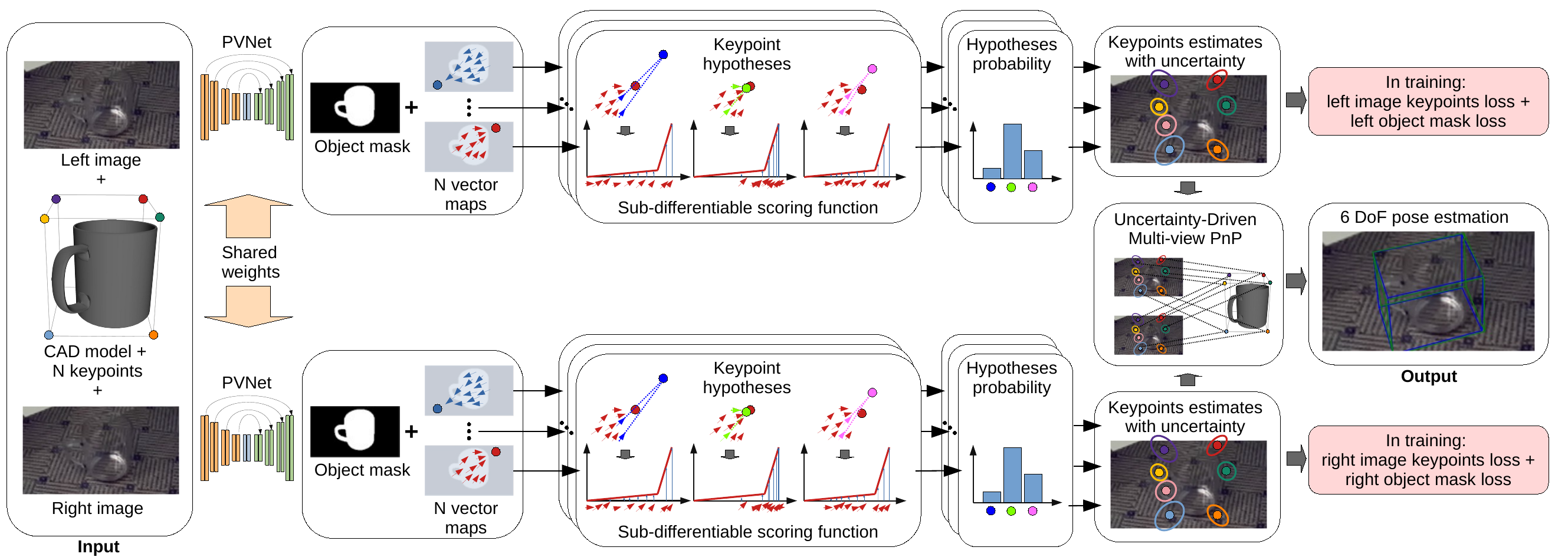}
   \caption{Overview of the KVN pipeline: i) the input stereo pair is processed by a shared-weights PVNet architecture to obtain an object segmentation mask and pixel-wise vectors pointing to each keypoints' projections; ii) for each keypoint, a set of $N_h$ hypotheses is obtained via minimal set sampling (here  $N_h=3$), then each hypothesis is weighted by all vectors by using our sub-differentiable scoring function; iii) for each keypoint, we compute a probability distribution over all valid hypotheses; iv) from each distribution, a keypoint is estimated along with its uncertainty: the position of the keypoints and the object mask are optimized during training; v) an uncertainty-driven multi-view PnP module (UM-PnP) estimates the object pose minimizing stereo reprojection errors weighted on the inverted keypoint covariance.}
   \label{fig:concept}
   \vspace{-0.5cm}
\end{figure*}
Object pose estimation methods can be classified by the number of stages they include: two-stage frameworks (e.g.,  \cite{9309178,9010850,Yu20206DoFOP}) use deep networks to match a set of 3D model points to the corresponding 2D projections in the image plane, and then use a second stage to estimate the 6D pose of objects from the correspondences, i.e, by using robust geometry-based algorithms such as variants of the RANSAC+PnP combination; single-stage frameworks (e.g., \cite{xiang2018posecnn,9156435}) directly regress from deep networks the 6D poses.
Compared to the latter, the former are not end-to-end trainable while deep networks usually rely on surrogate loss (for example, point-to-point distance) which does not directly reflect the quality of the final 6D pose estimate. Despite that, we believe two-stage approaches still remain attractive for many reasons, among others: i) when the 2D-3D correspondences are accurate, a combination of RANSAC + PnP still provides superior results \cite{9156435}; ii) they are generally more flexible when dealing with multi-view pose estimation tasks since it is possible to directly apply well-established multi-view geometry techniques \cite{9636459}.\\
The non-differentiable nature of the RANSAC + PnP combination forces some approaches to be trained only on non-optimal surrogate losses defined on intermediate pose representations \cite{9309178,Yu20206DoFOP}, while other methods design ad-hoc neural networks to mimic the behavior of the desired algorithm \cite{Song2020HybridPose6O,JIANG202216}. Differentiable variants of RANSAC that allow gradient flow while retaining most of its original formulation have been designed in other domains, e.g. for camera localization \cite{8099750,8578587,wei2022fully}, while \cite{campbell2020solving,9156614} back-propagate through a standard PnP optimizer by leveraging the implicit function theorem.\\

In this work, we propose KVN (Keypoints Voting Network, \figref{fig:concept}), a novel stereo pose estimation pipeline that fuses information from parallel monocular networks with an Uncertainty-driven Multi-view PnP optimizer (UM-PnP). Inspired by \cite{8099750,8578587}, we integrated PVNet \cite{9309178}, an established monocular pose estimation network, with a differentiable RANSAC layer that enables the network to directly infer 2D-3D keypoint correspondences.   
Correspondences extracted from each image acquired by the stereo rig are then fused inside the UM-PnP solver to estimate the objects' pose.  
\subsection{Contributions}
Our contributions are the following:
i) A novel 6D stereo object pose estimation pipeline that extends PVNet \cite{9309178} with a differentiable RANSAC layer, turning it into a keypoint prediction network without losing the robustness granted by its pixel-wise voting approach; ii) a novel sub-differentiable hypotheses' scoring function along with an ablation study that supports our choice over previous proposals; iii) a challenging, real-world and fully annotated stereo object pose estimation dataset (TTD: Transparent Tableware Dataset\footnote{Available at: \url{https://doi.org/10.5281/zenodo.10580443}}); iv) an extensive performance evaluation on the challenging task of transparent object pose estimation, showing that our method achieves state-of-the-art performance on both the public TOD dataset \cite{liu2020keypose} and TTD; v) An open-source implementation of KVN\footnote{Available at: \url{https://github.com/ivano-donadi/KVN}}.
\section{Related Work}\label{sec:rel_work}
This section showcases significant research contributions that focus on monocular, stereo, and RGB-D-based object pose estimation. Due to the extensive research interest in monocular pose estimation, we further categorize these approaches into three sub-groups \cite{Fan2022}: direct methods that perform end-to-end pose regression, keypoint-based methods that predict sparse 2D-3D keypoint correspondences and use PnP to derive the pose, and dense coordinate-based methods that operate as keypoint-based approaches but rely on dense per-pixel 2D-3D correspondences.

\subsubsection{Direct Methods}

Direct methods have the advantage of being end-to-end trainable with losses formulated directly on the pose but generally lack the capability of exploiting explicit geometric constraints.
PoseCNN \cite{xiang2018posecnn} uses Hough voting to aggregate pixel votes for object centroid 2D position and depth, and image features to predict orientation. EfficientPose \cite{Bukschat2020EfficientPoseAE}, extends an existing architecture for object detection \cite{9156454} with rotation and translation sub-networks. SSD-6D \cite{8237431} predicts viewpoint, in-plane rotation, and confidence for each object detected and uses projective geometry to obtain a 6D pose hypothesis. GDRNet \cite{9578682} predicts object segmentation, 2D-3D correspondences, and object surface region classification, which are used to predict the pose using a convolutional PnP module. Finally, PPN \cite{9423353} introduces a pose refinement module that improves the accuracy of the estimation by aligning views produced by a differentiable renderer with the input image.

\subsubsection{Keypoint-based Methods}

Keypoint-based methods can take full advantage of the explicit geometric constraints between a set of 3D object keypoints and their 2D projection in the input image, but often rely on non-differentiable geometric algorithms such as RANSAC and PnP to obtain the pose, in which case the network cannot be trained with a loss formulated directly on the pose. 
BB8 \cite{8237675} was a pioneer in this field, and used a convolutional network to perform object segmentation and predict the 2D projections of the eight bounding box corners around the object. PVNet \cite{9309178} improved this method by predicting a unit vector pointing to each keypoint projection for each pixel in the image corresponding to the target object, and in \cite{Yu20206DoFOP} this approach is improved by accounting for the distance between pixels and keypoints. HybridPose \cite{Song2020HybridPose6O} extended PVNet by predicting additional geometric cues from the input image (such as edges and symmetry relations) and introducing a trainable module to optimize the object pose. MLFNet \cite{JIANG202216} predicts dense surface normals and 2D-3D correspondences as 3D intermediate geometric representations, used to compute the same 2D representations as HybridPose. \cite{Oberweger2018} regresses 2D heatmaps of object keypoints projections, and RTM3D \cite{rtm3d} improves the heatmap optimization result by inferring additional geometric cues, such as object size and centroid depth. YOLOPose \cite{yolopose} uses a transformer architecture and a learnable rotation estimation module to predict object rotation and translation. DGECN  \cite{9879658} introduces a learnable PnP network based on edge convolution to replace RANSAC+PnP.

\subsubsection{Dense Coordinate-based Methods}

This category of monocular pose estimator aims to make indirect pose estimation more robust against occlusions and cluttering by predicting a dense map of 2D-3D correspondences for each object pixel in the input image, providing a more rigid set of constraints to the subsequent PnP module. CDPN \cite{9009519}, Pix2Pose \cite{9008819}, DPOD \cite{9010850}, and EPOS \cite{9157391} all fall inside this category. Pix2Pose also adds dense predictions for occluded object parts and a GAN discriminator to distinguish between ground truth and predicted pose maps. DPOD textures dataset objects with a UV map and formulates correspondences prediction as UV color regression and, finally, EPOS subdivides the object model into surface fragments and estimates the dense 2D-3D correspondence map with respect to the center of each corresponding surface fragment.


\subsubsection{Differentiable Geometric Methods}

The main drawback of most indirect methods is that they include non-differentiable geometric algorithms into the pipeline (e.g. RANSAC and PnP), making the system not trainable in an end-to-end fashion. To address this issue,
DSAC \cite{8099750}, \cite{8578587} replaces the deterministic hypothesis selection of RANSAC with a probabilistic counterpart using a differentiable module that assigns scores to each hypothesis and computes their probability distribution using softmax; the winning RANSAC hypothesis is obtained by sampling this distribution. NGRANSAC \cite{9008398} extends DSAC by introducing a neural network that predicts a probability distribution over all data points for optimal minimal set sampling. $\nabla$-RANSAC \cite{wei2022fully} uses Gumbel Softmax Sampling with the straight through trick \cite{oord_2017} to allow for dense gradients in sampling operations. \cite{campbell2020solving} and \cite{9156614} treat PnP (Perspective-n-Point) solvers as black-box optimization blocks and differentiate its output with respect to its input via the implicit function theorem \cite{MingariScarpello2002}, while EPro-PnP \cite{9879345} proposes a novel probabilistic PnP layer predicting a distribution of pose in the SE(3) manifold.

\subsubsection{Stereo Object Pose Estimation}

The advantage of stereo object pose estimation methods is that they can learn to infer depth from pixel displacement between the two input views, solving the ambiguity problem of monocular pose estimation.
KeyPose \cite{liu2020keypose} generates 2D location and disparity heatmaps from a stack of stereo images, allowing for back projection to determine keypoint position and depth. GhostPose \cite{9636459} uses multi-view geometry to triangulate the 3D position of bounding box corners from their predicted projections, leveraging a canonical orientation for symmetric objects.
StereoPose \cite{chen2022stereopose} deals with transparent objects by using normalized coordinate space \cite{Wang_2019_CVPR} and left-right correspondences from both front and back views of the transparent objects.


\subsubsection{RGB-D Based Methods}

DenseFusion \cite{8953386} uses a dedicated network to fuse pixel-wise depth and color features to predict pose, which is shown to outperform simple feature concatenation. PointPoseNet \cite{9093272} predicts, for each point in the point cloud extracted from the RGB-D image, a 3D vector pointing to each object keypoint, and aggregates them to extract the final 3D keypoint position. RCVPose \cite{Wu2022} regresses the distance from each object point in the point cloud to each keypoint and obtains hypotheses by intersecting the spheres defined by object point and distance. \cite{9197461} uses two separate networks to predict translation and rotation from a semantically segmented input point cloud and refines the pose estimation using a geometry-based optimization process. ClearGrasp \cite{9197518} has been designed to deal with the challenges of transparent objects, where surface normals, masks, and occlusion boundaries extracted from RGB images are used to refine the initial depth estimates.
\subsubsection{Pose Estimation of Transparent Objects}
The problem of estimating transparent objects is relatively recent and has traditionally been addressed by designing ad-hoc edge detectors applied to RGB-D \cite{6577942} and stereo \cite{PhillipsRSS2016} setups. Among the deep learning-based approaches presented above, KeyPose \cite{liu2020keypose}, ClearGrasp \cite{9197518}, StereoPose \cite{chen2022stereopose}, and GhostPose \cite{9636459} were specifically designed to localize transparent objects.

\section{KVN: Keypoints Voting Network}\label{sec:method}

In this section, we first introduce PVNet, the pose estimation network our work is based on, and how we adapted it to the stereo domain by adding an uncertainty-driven multi-view PnP optimizer. Next, we present our architecture (KVN), the differentiable RANSAC layer we added to PVNet, our choice of losses, our training strategy, and the essential implementation details.
\subsection{PVNet: Pixel-Wise Voting Network}
\label{sec:pvnet}
Our work builds upon PVNet \cite{9309178}, a keypoint-based monocular pose estimation network. Given a CAD model of the object of interest, a set of 3D keypoints defined on the model coordinate system, and an RGB image $\mathcal{I}$, PVNet predicts an object segmentation mask and pixel-wise unit vectors pointing to each keypoint projection in the image (see \figref{fig:concept}). Votes from pixels that belong to the segmentation mask are aggregated using RANSAC, which returns the predicted keypoints position and their variance. An uncertainty-driven PnP module is then used to predict the object's pose by minimizing the reprojection errors of the keypoints in the image, weighted by the inverse of their variance. The advantages of this approach are that it is robust against cluttering and can infer the position of non-visible keypoints due to its underlying pixel-wise directional voting scheme.
Since RANSAC and PnP are non-differentiable algorithms, the network is trained with a smooth loss $l_1$ between predicted and ground truth voting vectors, on top of a cross-entropy segmentation loss. Usually, PVNet leverages 8 object keypoints selected with the Farthest Point Sampling (FPS) strategy, plus the object centroid.
The main drawback of this method is that the network does not necessarily converge to the solution maximizing pose accuracy, but rather to the one minimizing the voting error, which may not be the same. 

\subsection{UM-PnP: Uncertainty-driven Multi-view PnP}\label{sec:mv_pnp}
Given a multi-view stereo rig composed of a set of $M$ calibrated cameras that frame a scene containing objects of interest from several viewpoints, PVNet can be used independently for each image $\mathcal{I}_i$, $ i=0,1,\dots,M-1$, to infer the projections of the keypoints of the objects. We extend to this setup the Uncertainty-driven PnP algorithm presented in \cite{9309178}. Let $\Pi_i$ be the cameras perspective projection functions that map 3D scene points to image points, and $\{ (\mathbf{R}_i,\mathbf{t}_i)~~|~~\mathbf{R}_i \in SO(3),~ \mathbf{t}_i \in \mathbb{R}^3 \}$ the transformations that define the change of coordinates from the coordinate system of the stereo rig to the reference system of each camera. Without loss of generality the stereo rig frame is usually set on the frame of a camera, e.g., the first camera, $(\mathbf{R}_0,\mathbf{t}_0) = (\mathbf{I}, [0~0~0]^\top])$. Given a set of $N$ 3D object keypoints, $\{K_0,K_1, \dots ,K_{N-1}\}$ and, for each $i$-th image, the estimates of the 2D projections of such keypoints $\{k_{i,0}, k_{i,1},  \dots , k_{i,N-1}\}$, it is possible to estimate the object position $(\mathbf{R}_{obj},\mathbf{t}_{obj})$ with respect to the stereo rig frame by minimizing the squared reprojection error on all images as:
\begin{equation} \label{eq:sq_optimizer}
 \argmin_{\mathbf{R}_{obj},\mathbf{t}_{obj}}  \sum_{i=0}^{M - 1} \sum_{j=0}^{N - 1} \mathbf{e}_{i,j}^{\top}\mathbf{e}_{i,j}
\end{equation}
where:
\begin{equation}\label{eq:residual}
\mathbf{e}_{i,j} = \Pi_i\left( 
  \mathbf{R}_i(\mathbf{R}_{obj} K_j+\mathbf{t}_{obj})+ \mathbf{t}_i \right)- k_{i,j}
\end{equation}
is the two-dimensional residual.

This approach applied for a stereo rig to the result of two independent PVNet predictions was proposed in \cite{Liu2021StereOBJ1MLS} under the name of ``object triangulation'': it allows the PnP solver to leverage twice the number of 2D-3D correspondences compared to the monocular version to optimize the same number of parameters. We further evolve this optimizer by weighting each reprojection by the keypoint's variance returned by RANSAC. Let $\Sigma_{i,j}$ be the 2x2 covariance matrix of keypoint $j$ in the $i$-th image computed as described in \secref{sec:imp_det}, we minimize the following cost function:
\begin{equation} \label{eq:sq_w_optimizer}
 \argmin_{\mathbf{R}_{obj},\mathbf{t}_{obj}}  \sum_{i=0}^{M - 1} \sum_{j=0}^{N - 1}  \mathbf{e}_{i,j}^T \Sigma_{i,j}^{-1}\mathbf{e}_{i,j}
\end{equation}
We refer to this solver as UM-PnP.
\subsection{Differentiable RANSAC Voting Scheme}\label{sec:dsac}
To estimate the objects' poses, the multi-view PnP algorithm presented in \secref{sec:mv_pnp} requires for each $i$-th view the 2D keypoints projections $k_{i,j}$, whereas the standard PVNet framework is trained only on a surrogate of keypoints, i.e., a set of vectors pointing to keypoints. In this section, we describe the differentiable RANSAC (DSAC) layer we designed, inspired by \cite{8099750, 8578587}, to make PVNet end-to-end trainable directly on 2D keypoints projections. 
For each image $\mathcal{I}_i$, the output of the standard PVNet is (see \figref{fig:concept}):
\begin{itemize}
    \item a segmentation mask $M(p)\in[0,1]$ which, for each image pixel $p \in \Omega$, with $\Omega$ the image domain, provides non-zero values only for pixels belonging to the projection of the object of interest;
    \item for each keypoint $K_j$, a pixel-wise vector map $V_j(p) \in \mathbb{R}^2$ that holds the vectors pointing to the projection of the keypoint. 
\end{itemize}
Following the classical RANSAC pipeline, it is possible to generate a pool of $N_h$ hypotheses $\{h_{j,0},h_{j,1}, \dots, h_{j,N_h-1}\}$ for each keypoint $K_j$ via minimal set sampling. In PVNet's case, minimal sets contain two non-parallel unit vectors, uniformly sampled from $V_j(p')$, with $\{p' \in \Omega| M(p') \neq 0 \}$. Given two unit vectors $v_r, v_s \in V_j(p')$, centered in pixels $p_r$ and $p_s$ respectively, the related hypothesis $h_{j,l}$, $l = 0,,\dots,N_h-1$, should satisfy the following equality:
\begin{equation}
h_{j,l} = p_r+av_r=p_s+bv_s
\label{eq:hyp_eq}
\end{equation}
with  $a$, $b \in \mathbb{R}$. \eqref{eq:hyp_eq} can be solved by using the perp-dot-product \cite{hill1994pleasures}, which is differentiable under the condition that the two vectors are not parallel:
\begin{equation}
    a = \frac{(p_s-p_r)\bot v_s}{v_r\bot v_s} 
\end{equation}
where $\bot$ stands for the perp-dot-product operator, $u \bot w = u_xw_y - u_yw_x = |u|~|v|~sin\theta$, with $\theta$ the angle from vector $u$ to vector $w$. 
If the two vectors are parallel, the hypothesis is classified as invalid hence its gradients will not be computed. The next step would be counting the inliers for hypothesis $h_{j,l}$. PVNet classifies a pixel $p$ as inlier if the cosine similarity between the estimated unit vector $v$ and the ground truth vector connecting $p$ to the projection of $K_j$ is greater or equal than a fixed threshold, usually 0.99. However, this step is not differentiable so, inspired by \cite{8578587}, we replace the inlier count with a summation of individual pixels' scores. Assuming $f$ is a differentiable or sub-differentiable function in the interval $[-1;1]$ and $S_C$ is the cosine similarity operator, the inlier score for hypothesis $h_{j,l}$ is defined as:
\begin{equation}\label{eq:sim_score}
    S_I(h_{j,l}) = \sum_{p \in \Omega} M(p)f(S_C(V_j(p), h_{j,l}-p)) 
\end{equation}

It is possible, at this point, to define a probability distribution for all valid hypotheses in the pool, i.e. hypotheses generated by non-parallel unit vectors, using the softmax operator:
\begin{equation}\label{eq:softmax}
    P(h_{j,l}) = \frac{ \text{exp} (S_I(h_{j,l})) }{ \sum\limits^{N_h}_{\substack{l^{'}=0\\h_{j,l^{'}} \text{ valid}}}\text{exp} (S_I(h_{j,l^{'}})) }
\end{equation}
\subsection{Cost Function and Training}

\eqref{eq:sim_score} requires a function $f$ remapping the cosine similarity $s = S_C(\cdot)$ to an inlier score. In standard RANSAC, $f$ is a Heaviside function centered at the RANSAC threshold $t$. We replace it with a sub-differentiable leaky-ReLU inspired piecewise linear function, parametrized by the RANSAC inlier threshold $t$ and the ``soft-inlier'' value at the threshold ($v>0$):
\begin{equation} \label{eq:sub_diff_func}
f(s; t,v) = \begin{cases} 
 \frac{1-v}{1-t}(s-1)+1 & s \ge t \wedge s \le 1\\
 \frac{v}{t+1}(s+1) & s \ge-1 \wedge s < t
\end{cases} 
\end{equation}
An example of this function, with $t=0.9$ and $v=0.1$ can be seen in \figref{fig:activations} (red plot). 
Using this scoring function, all involved pixels (see \eqref{eq:sim_score}) will cast a non-zero vote for each hypothesis, resulting in non-vanishing gradients even for alleged outliers, so enabling a smooth optimization. 
We train our model by minimizing the following loss:
\begin{equation}\label{eq:kvnet}
l_{KVN} = l_{mask} + l_{DSAC}
\end{equation}
that includes a cross-entropy loss $l_{mask}$ for the segmentation mask and a keypoints loss that takes into account the average squared keypoint error over the hypotheses distribution:
\begin{equation}\label{eq:disac}
    l_{DSAC} = \sum_{j=0}^N~\sum_{\substack{l=0 \\ h_{j,l} \text{ valid}}}^{N_h} P(h_{j,l}) \cdot  ||h_{j,l}-k_j^*||^2
\end{equation}
While hypotheses generation is performed only on a subset of pixels pairs, \emph{all} object's pixels are involved in \eqref{eq:softmax} hence in \eqref{eq:disac}, so probability gradients back-propagate through each object pixel. To stabilize training, we control the broadness of the hypotheses distribution with a learnable softmax temperature $1/\alpha$. Parameter $\alpha$ is trained with an $l_1$ loss on the distribution's entropy, as in \cite{8578587}. This trick avoids degenerate hypothesis distributions that would not take advantage of the full range of RANSAC rounds.
\subsection{Implementation Details}
\label{sec:imp_det}
We train KVN on batches composed of stereo pairs, i.e., couples of images composed by the left and the corresponding right view.
This means that the training pipeline is actually monocular, but batches always include both images for each stereo pair. During inference, differentiable RANSAC is replaced by standard RANSAC, granting the same inference speed as classical PVNet. 

In the very first training epoch, votes for each pixel are assigned semi-randomly and can lead to keypoint hypotheses far outside image boundaries, resulting in large task losses. In this scenario, the network converges either very slowly or not at all. 
To avoid this situation, we prune hypotheses with an $x$ or $y$ coordinate located outside the image by a margin greater than the corresponding image dimension.

In all our experiments, we select the best keypoint hypothesis using 256 RANSAC rounds (128 for the TTD dataset), and the corresponding variance is computed (at inference time only) with respect to a set of 1024 additional RANSAC hypotheses, each weighted by its inlier count.
We set the target entropy for the hypotheses distribution to 6, following \cite{8578587}. The network is trained for 150 epochs using an ADAM optimizer with a learning rate of 1e$^{-3}$ decayed by $\gamma=0.5$ every 20 epochs. We perform data augmentation by random background with 50\% probability, random $x$-axis tilting between -10$\degree$ and 10$\degree$ and the stereo mirroring operation described in \cite{liu2020keypose} with 50\% probability. We adopt the same photometric augmentation strategy of KeyPose \cite{liu2020keypose} and normalize images using ImageNet \cite{deng2009imagenet} statistics.

\section{Experiments}\label{sec:experiments}

\vspace{-0.1cm}
\subsection{Datasets}
\begin{figure}[t]
\centering
\begin{minipage}[b]{\columnwidth}
  \centering
  \includegraphics[width=.49\linewidth]{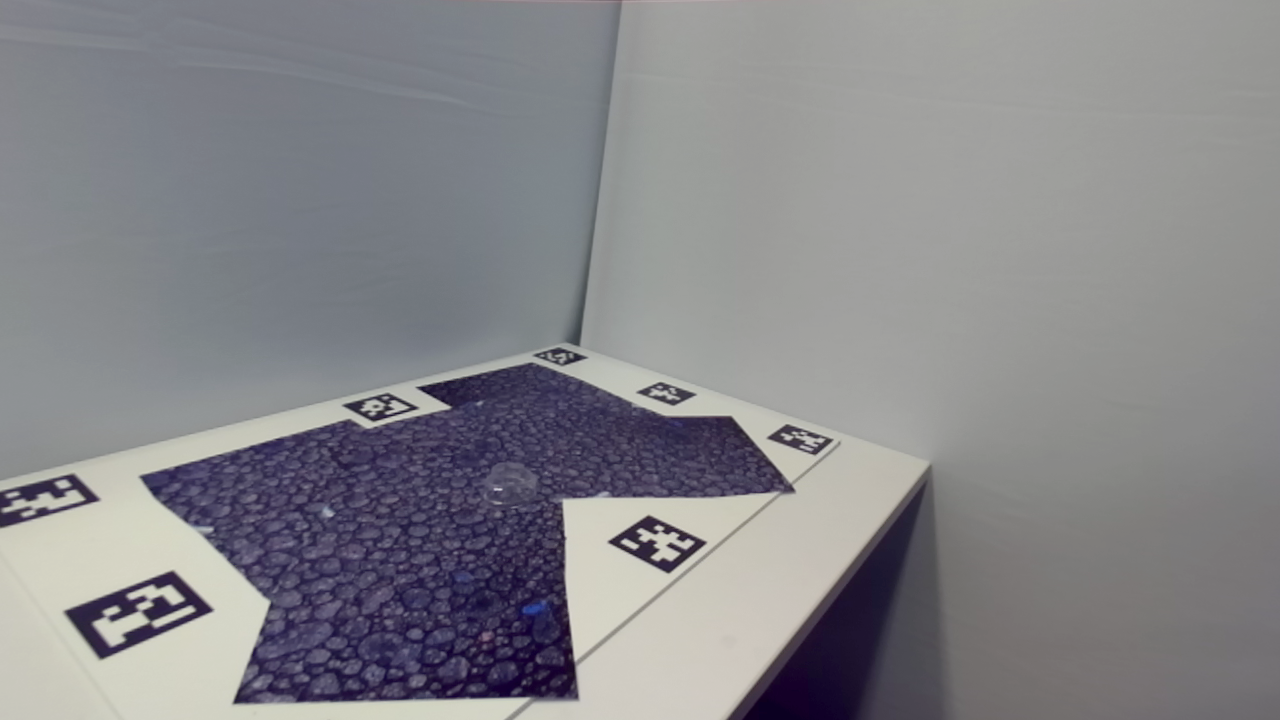}
  \includegraphics[width=.49\linewidth]{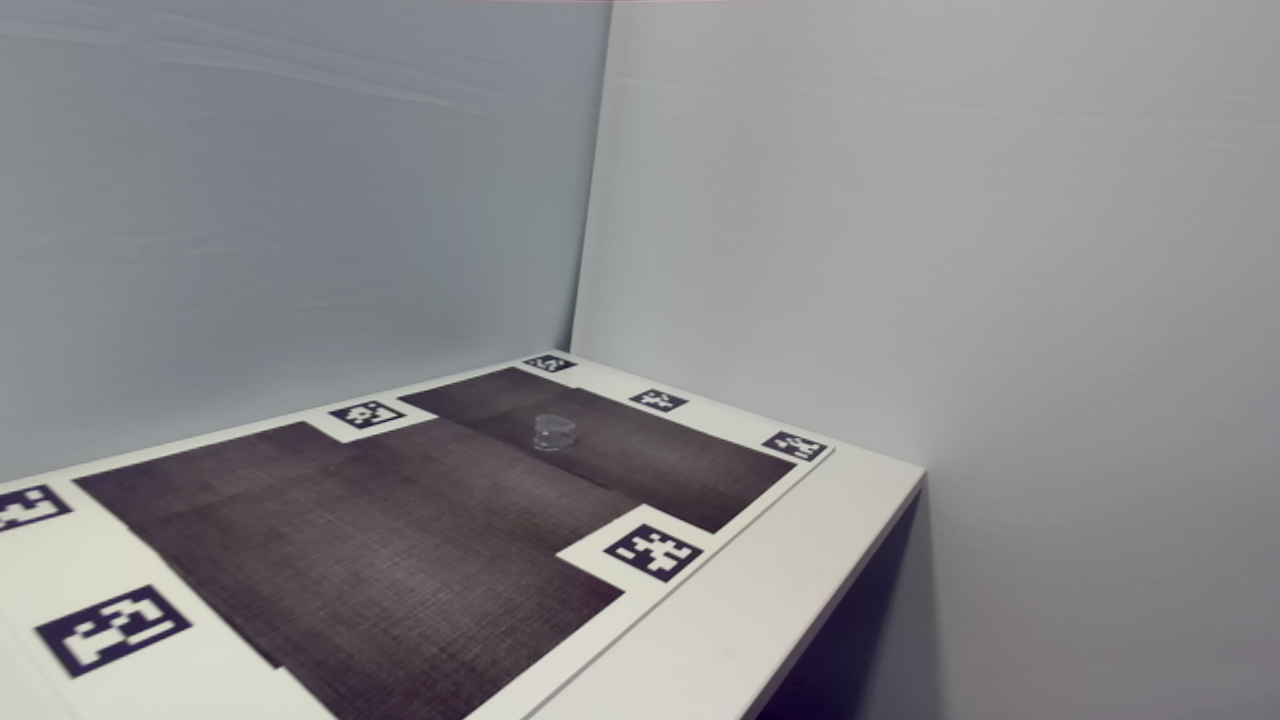}
\end{minipage}\\
\vspace{1mm}
\begin{minipage}[b]{\columnwidth}
  \centering
  \includegraphics[width=.49\linewidth, height=3.1cm]{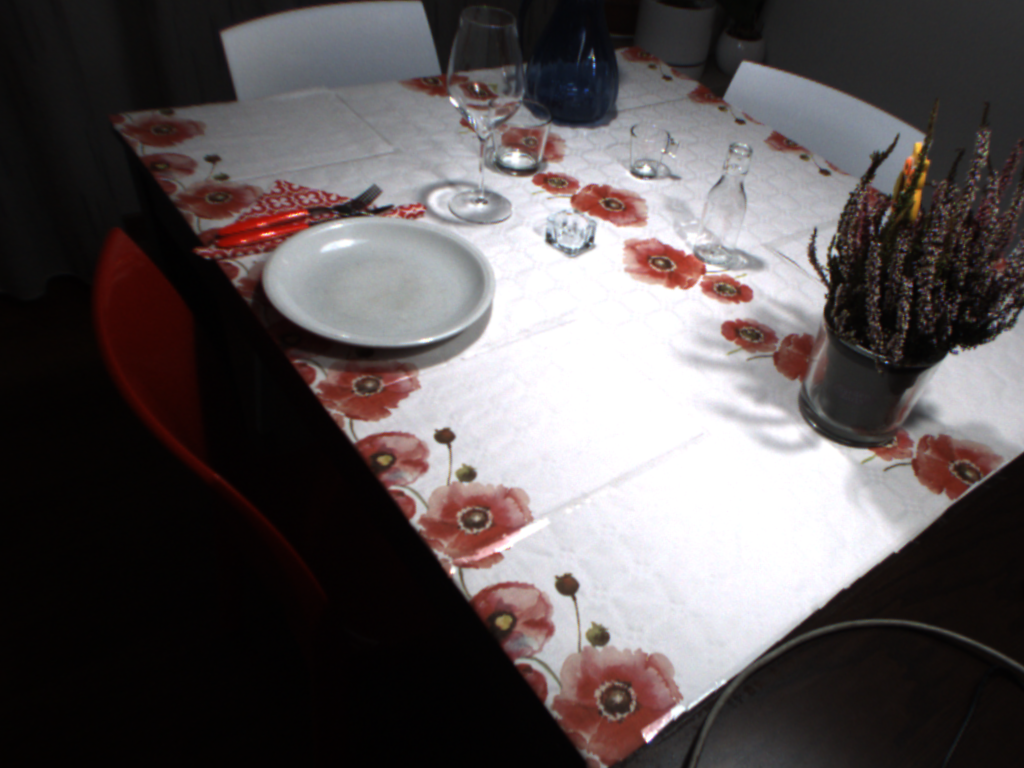}
  \includegraphics[width=.49\linewidth, height=3.1cm]{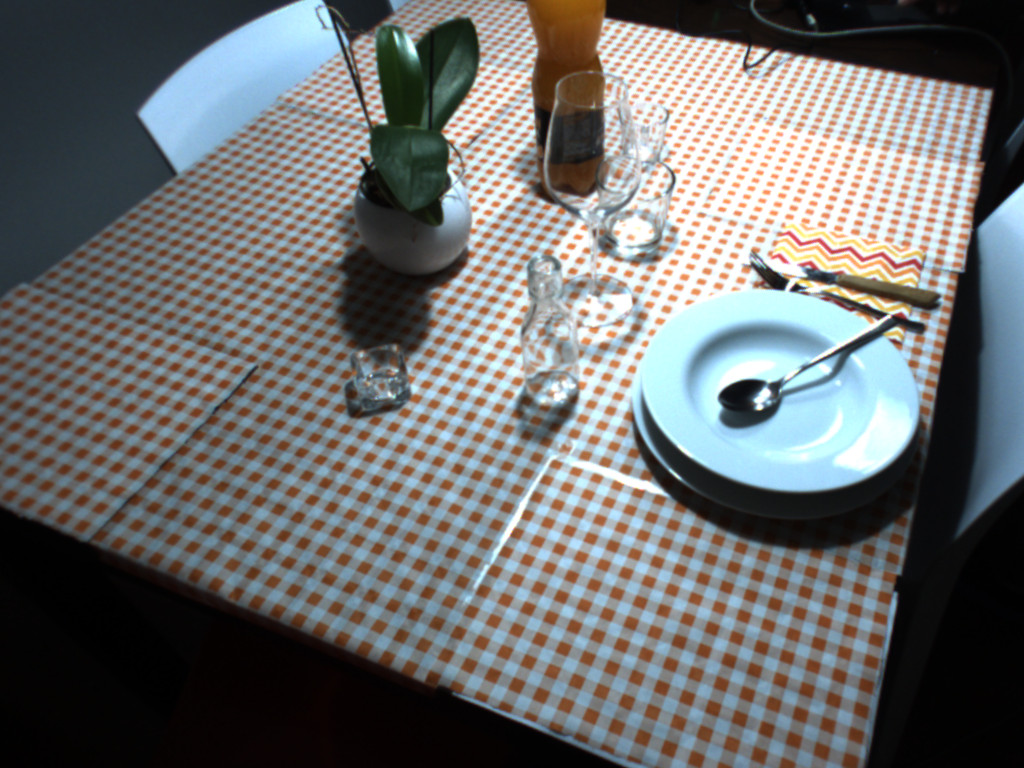}
\end{minipage}
\caption{(Top row) The heart$_0$ object from the TOD dataset over two differently textured mats; (Bottom row) Two images from the TTD dataset.
}
\label{fig:tod_ttd}
\vspace{-0.6cm}
\end{figure}
Our approach is evaluated on two datasets:\\
\textbf{Transparent Object Dataset (TOD)} \cite{liu2020keypose} for stereo object pose estimation, including about 45,000 stereo images of 15 different objects, 
along with the ground truth segmentation masks for the left camera, pose annotations, and depth maps. 
For each object, the stereo images are organized into 10 sub-datasets, each obtained by placing the object on a differently textured mat, as shown in \figref{fig:tod_ttd} (top row). The standard evaluation procedure defined in \cite{liu2020keypose} is to select 9 out of 10 textured mats for training, and then test on the remaining textured mat.\\
\textbf{Transparent Tableware Dataset (TTD)}, a newly created dataset made available with this paper that includes 600 stereo images of real scenes (e.g., \figref{fig:tod_ttd} (bottom row)) with some objects positioned in the scene, including 5 transparent objects. For each transparent object the CAD model, pose annotations and segmentation masks are provided. Unlike TOD, in TTD there are several occlusions because all transparent objects together with others are present in the scene at the same time while the fiducial markers used for pose annotation are not visible, thus reducing any perceptual bias. Finally, the limited number of images makes it more similar to datasets that can be acquired for real-world applications. For each object in the dataset, the standard "mixed" benchmark has been tackled, where  60\% of the available images are used for training, 20\% for validation, while evaluation is performed on the remaining 20\%. We performed 6 independent training processes for each method, reporting for each object the average results obtained on the test set.
\vspace{-2mm}
\subsection{Metrics}


Given the ground truth and predicted object pose, and the set of 3D keypoints defined on the object's CAD model we can compute three well-known pose estimation metrics with standard thresholds:

\begin{itemize}
    
    \item $<$2cm: the percentage of test samples for which the average distance between ground truth and estimated 3D keypoint positions is less than 2 cm. If the object is symmetric, the ambiguity in the keypoints association is solved by matching each estimated keypoint with its closest ground truth counterpart;
    \item ADD(-S): the percentage of test samples for which the average distance between ground truth and estimated 3D keypoint positions is less than 10\% of the object's diameter. If the object is symmetric, the ambiguity in the keypoints association is solved by pairing each estimated keypoint with its closest ground truth counterpart;
    \item Area Under Curve (AUC): area under a curve in which each point represents the ADD(-S) metric obtained using the corresponding x-axis value as a threshold, in our case from 0 cm to 10 cm. 
    \item Mean Absolute Error (MAE): the average absolute 3D position error between ground truth and estimated keypoints (in mm);
\end{itemize}
In calculating the aforementioned metrics, we used the set of 3D keypoints defined in the TOD dataset for each object.

\subsection{Scoring Function Evaluation}

\begin{figure}[t!]
   \begin{center}
   \includegraphics[width=.8\linewidth, height=40mm]{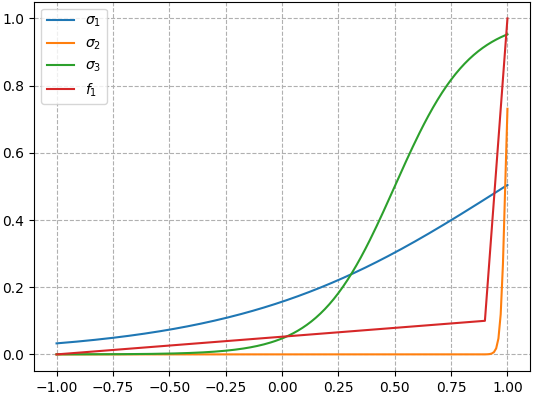}
   \end{center}
   \vspace{-2mm}
   \caption{Four scoring functions compared in the evaluation of differentiable RANSAC. The x-axis represents the cosine similarity between estimated and ground truth unit vectors.}
   \label{fig:activations}
\end{figure}

In \cite{8578587}, the authors replaced the non-differentiable inlier-counting step of RANSAC with a sum of ``soft-inlier'' values, obtained by remapping similarity scores to the $[0,1]$ range with a sigmoid scoring function parametrized by its center ($\tau$) and its slope ($\beta$):

\begin{equation}
    \sum_{i} \sigma(s_i;\beta, \tau)  = \sum_{i} sig(\beta(s_i - \tau))
\end{equation}

where $sig(*)$ is the sigmoid function and $s_i$ is a similarity measure between the current RANSAC hypothesis and a data point $i$. 
In this section, we support the choice of our sub-differentiable scoring function $f$, defined in \eqref{eq:sub_diff_func}, by comparing it with three differently parametrized sigmoids (\figref{fig:activations}). 
We have selected the sigmoids which empirically obtained the best results in our experiments ($\sigma _1$ and $\sigma_2$): both share the same threshold $\tau$, while surprisingly the parameter $\beta$, which determines a strong difference in shape between the two, seems to have little influence. For completeness, we have also reported a sigmoid ($\sigma_3$) with a lower $\tau$ value.
\tabref{tab:ransac_eval} shows the average values of MAE, AUC, and $<$2cm metrics for the four different models on a representative for each of the three object categories present in the dataset: mirror symmetrical (heart$_0$), asymmetrical (mug$_6$) and symmetrical w.r.t. an axis (cup$_1$). 
The proposed scoring function $f$ slightly outperforms the others.
Furthermore, it is more intuitive to choose the parameters for $f$ than for the sigmoid, since $t$ can be set equal to the standard RANSAC threshold while  $v$ models the maximum voting power given to outliers. The sigmoid's center $\tau$ cannot be directly related to the RANSAC threshold without choosing a very steep slope $\beta$ too, but in this case the gradients explode near the maximum value of the cosine similarity, as in $\sigma _2$; conversely, with small $\beta$ the range is limited to a maximum value close to $0.5$, as in $\sigma _1$.
We also tried to let the network learn optimal sigmoid values for $\beta$ and $\tau$; however, the resulting sigmoid converges to the step function centered near $x=1$, actually preventing the network from learning how to estimate keypoints.

\begin{table}[t]
\vspace{-2mm}
\centering
\scriptsize
\begin{tabular}{|c|c c c|}
     \hline
     scoring function & $<$2cm & MAE & AUC \\
     \hline
     $\sigma_1(*;\beta=1.7,\tau=0.99)$ & 91.3 & 9.9 & 90.8\\ 
     $\sigma_2(*;\beta=100,\tau=0.99)$ & 91.1 & 9.9 & 90.8 \\
     $\sigma_3(*;\beta=6,\tau=0.5)$ & 90.2 & 10.8 & 90.1 \\
     $f(*;t=0.9,v=0.1)$ & \textbf{91.7} & \textbf{9.8} & \textbf{90.9} \\ 
     \hline
\end{tabular}
\caption{Evaluation of scoring functions for differentiable RANSAC.}
\label{tab:ransac_eval}
\vspace{-0.5cm}
\end{table}

\subsection{Comparison With the State-of-the-Art Methods}

\begin{table*}[t]
\resizebox{\textwidth}{!}{%
    \centering
        \begin{tabular}{|c|c|c c c c c c c c c c c c c c c |c|}
    \hline
        Method & Metric & ball$_0$          & bottle$_0$      & bottle$_1$      & bottle$_2$ & cup$_0$      & cup$_1$       & mug$_0$    & mug$_1$ & mug$_2$ & mug$_3$       & mug$_4$       & mug$_5$       & mug$_6$       & heart$_0$         & tree$_0$ & mean\\
    \hline
        \multirow{ 3}{*}{DenseFusion-O}  & AUC   & 90.0          & 88.6          & 69.1          & 56.0     & 84.0          & 80.7          & 67.8          & 66.3 & 71.4    & 70.0          & 69.0          & 76.8          & 51.2          & 61.7          & 75.5  & 71.9 \\
          & $<$2cm  & 94.4          & 97.8          & 9.1           & 28.4     & 79.1          & 65.3          & 12.5          & 10.3 & 28.1    & 20.3          & 4.7           & 41.9          & 3.1           & 17.2          & 50.9  & 37.5\\
          & MAE   & 9.9           & 11.3          & 57.6          & 77.8     & 16.0          & 37.5          & 32.2          & 33.7 & 28.6    & 30.0          & 31.0          & 23.2          & 75.2          & 38.3          & 24.5  & 35.1 \\
    \hline
        \multirow{ 3}{*}{DenseFusion-T}  & AUC   & 84.7          & 81.6          & 72.3          & 47.5     & 59.4          & 77.8          & 54.5          & 51.3 & 60.4    & 67.3          & 48.1          & 70.6          & 64.9          & 61.2          & 55.6  & 63.8 \\
          & $<$2cm  & 78.8          & 67.5          & 18.1          & 9.1      & 5.6           & 54.4          & 4.6           & 0.3  & 12.2    & 8.1           & 0.0           & 20.0          & 4.7           & 0.0           & 0.0   & 18.9 \\
          & MAE   & 15.3          & 18.4          & 27.6          & 65.6     & 40.5          & 22.1          & 45.5          & 48.7 & 39.5    & 32.7          & 54.9          & 29.4          & 35.9          & 38.8          & 44.4  & 37.2\\
    \hline
        \multirow{ 3}{*}{KeyPose}    & AUC       & 96.1 & \textbf{95.4} & \textbf{94.9} & 90.7 & 93.1 & 92.0 & 91.0 & 78.1 & 89.7 & 88.6 & 87.8 & \textbf{91.0} & \textbf{90.3} & 84.3 & 87.1 & 90.0\\
            & $<$2cm    & \textbf{100} & \textbf{99.8} & \textbf{99.7} & 91.4 & \textbf{97.8} & 95.3 & \textbf{94.6} & 63.6 & 90.1 & 87.2 & 87.1 & 93.1 & 92.2 & 77.2 & 82.5 & 90.1\\
            & MAE       & 3.8 & \textbf{4.6} & \textbf{5.1} & 9.3 & 6.8 & 7.1 & \textbf{8.9} & 21.9 & 10.1 & 11.3 & 12.1 & \textbf{9.0} & \textbf{9.7}  & 15.6 & 12.8 & 9.9\\
    \hline
    \multirow{ 3}{*}{GhostPose} & AUC       & -   & 94.5  & 92.4 & 90.2 & 92.7 & 91.8 & 90.1 & \textbf{90.0}  & 90.1 & 89.7 & 90.3 & 89.0 & 88.2 & - & - & -\\
     & $<$2cm    & -   & 96.4  & 95.2 & 91.6 & 92.6 & 94.0 & 93.4 & \textbf{94.8} & 93.4 & 92.9 & 95.8 & \textbf{93.6} & \textbf{93.1} & - & - & -\\
     & MAE       & -   & 5.3   & 7.8  & 9.4  & 7.2  & 8.4  & 10.1 & \textbf{10.8} & 10.2 & \textbf{10.1} & 9.0  & 10.6 & 11.5 & - & - & -\\ 
    \hline
        \multirow{ 3}{*}{s-PVNet} & AUC     & 96.6 & 94.7 & 93.7 & 89.9 & 93.1    & 94.3 & 85.8 & 79.7 & 88.4 & 87.3 & 89.4 & 86.0 & 87.2 & 89.5 & 90.5 & 89.7  \\
         & $<$2CM  & 97.7 & 95.2 & 96.6 & 91.1 & 96.8    & 97.1 & 84.4 & 71.9 & 88.7 & 88.5 & 91.8 & 90.3 & 89.5 & 85.5 & 92.8 & 90.5\\
         & MAE     & 3.9  & 5.8  & 6.7  & 10.7 & 7.5    & 6.1 & 15.8 & 21.7 & 12.8 & 14.7 & 11.6 & 16.3 & 14.1 & 10.9 & 10.1 & 11.2 \\
    \hline
        \multirow{ 3}{*}{KVN} & AUC        & \textbf{97.4} & 95.0 & 94.0 & \textbf{92.5} & \textbf{94.2} & \textbf{94.4} & \textbf{91.1} & 82.6 & \textbf{91.4} &\textbf{91.5} & \textbf{91.9} & 89.8 & 88.3 & \textbf{90.4} & \textbf{91.1} & \textbf{91.7}\\
        & $<$2CM     & 99.5 & 95.4 & 96.8 & \textbf{94.8} & 97.6 & \textbf{97.3} & 93.6 & 77.4 & \textbf{94.2} & \textbf{93.3} & \textbf{96.2} & 93.2 & 90.7 & \textbf{87.2} & \textbf{93.4} & \textbf{93.4} \\
        & MAE        & \textbf{3.1}  & 5.4 & 6.4  & \textbf{8.1}  & \textbf{6.3} & \textbf{6.0}  & 9.5  & 18.3  & \textbf{9.5}  & \textbf{10.1}          & \textbf{8.9}  & 12.0 & 13.2 & \textbf{10.0} & \textbf{9.4} & \textbf{9.1} \\
    \hline
    \end{tabular}}
    \caption[~Results summary]{Evaluation results on the TOD dataset. The metrics for each object are the mean of the results on the 10 different textures.
    }
    \label{tab:full-results}
    \vspace{-0.3cm}
\end{table*}

\begin{table}[t]
    \centering
    \scriptsize
    \begin{tabular}{|c|c|c c c c c|c|}
        \hline
        Method & Metric & w. glass & cup & holder & glass & bottle & mean \\
        \hline
        \multirow{ 2}{*}{s-PVNet} 
        & ADD(-S) & \textbf{95.6} &	30.6 &	14.5 &	82.8 &	84.2 &	61.5\\
        & $<$2cm & \textbf{93.9} & 68.0 & 73.0 & \textbf{92.5} & 87.9 & 83.1\\
        \hline
        \multirow{ 2}{*}{KVN} 
        & ADD(-S) &  92.4 & \textbf{33.2} & \textbf{18.9} & \textbf{83.7} & \textbf{85.1} & \textbf{62.7}\\
        & $<$2cm & 90.1 & \textbf{75.8} & \textbf{74.5} & 90.8 & \textbf{88.0} & \textbf{83.8}\\
        \hline
    \end{tabular}
    \caption{Evaluation results on the TTD dataset}
    \label{tab:TTD_results}
    \vspace{-8mm}
\end{table}

\begin{figure*}[t]
\centering
\begin{subfigure}{.39\linewidth}
  \centering
  \begin{subfigure}{.495\linewidth}
       \includegraphics[width=\linewidth,trim={0.5cm 2cm 0.5cm 2cm},clip]{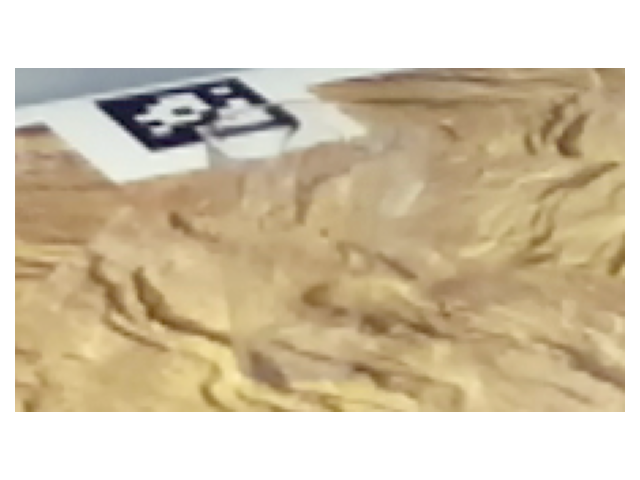}
  \end{subfigure}%
  \hspace*{\fill}
  \begin{subfigure}{.495\linewidth}
      \includegraphics[width=\linewidth,trim={0.5cm 2cm 0.5cm 2cm},clip]{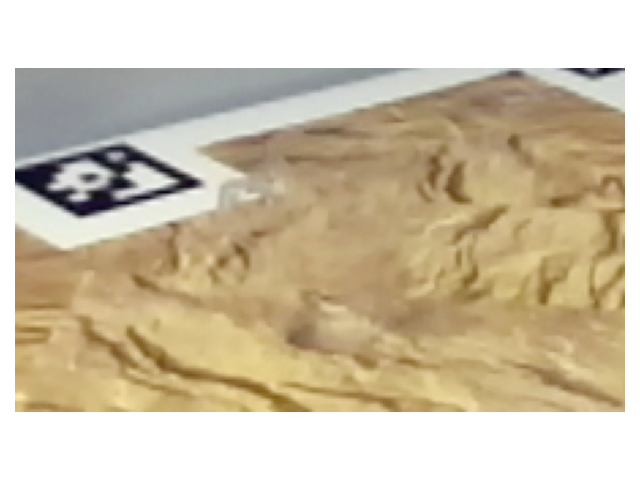}
  \end{subfigure}
  \par\smallskip
  \begin{subfigure}{0.495\linewidth}
      \includegraphics[width=\linewidth,trim={0.5cm 2cm 0.5cm 2cm},clip]{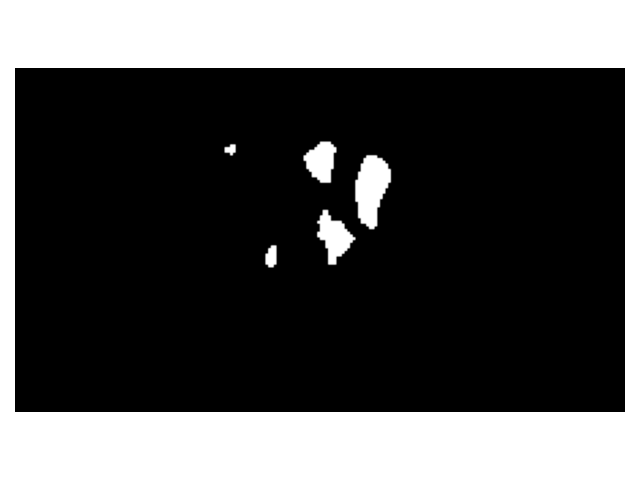}
  \end{subfigure}%
  \hspace*{\fill}
  \begin{subfigure}{0.495\linewidth}
      \includegraphics[width=\linewidth,trim={0.5cm 2cm 0.5cm 2cm},clip]{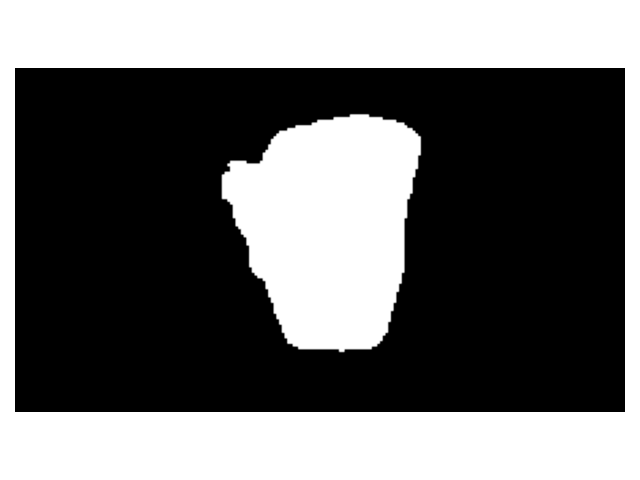}
  \end{subfigure}
  \par\smallskip
  \begin{subfigure}{0.495\linewidth}
      \includegraphics[width=\linewidth,trim={0.5cm 2cm 0.5cm 2cm},clip]{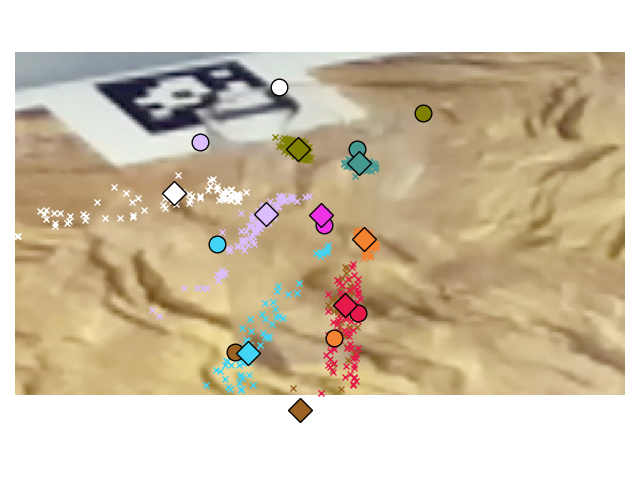}
  \end{subfigure}%
  \hspace*{\fill}
  \begin{subfigure}{0.495\linewidth}
      \includegraphics[width=\linewidth,trim={0.5cm 2cm 0.5cm 2cm},clip]{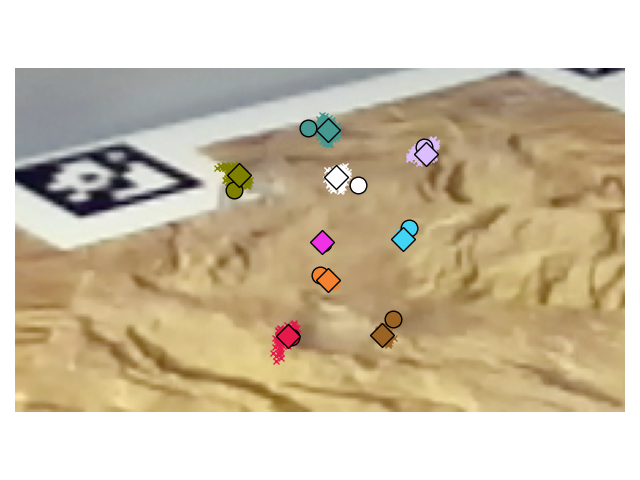}
  \end{subfigure}
  \par\smallskip
  \begin{subfigure}{0.495\linewidth}
      \includegraphics[width=\linewidth,trim={0.5cm 2cm 0.5cm 2cm},clip]{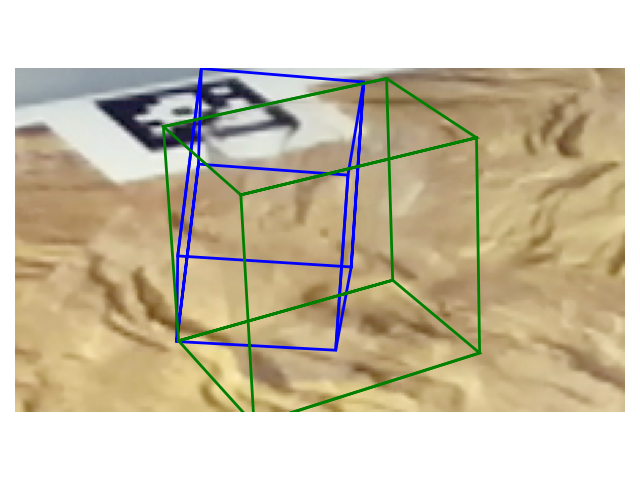}
  \end{subfigure}%
  \hspace*{\fill}
  \begin{subfigure}{0.495\linewidth}
      \includegraphics[width=\linewidth,trim={0.5cm 2cm 0.5cm 2cm},clip]{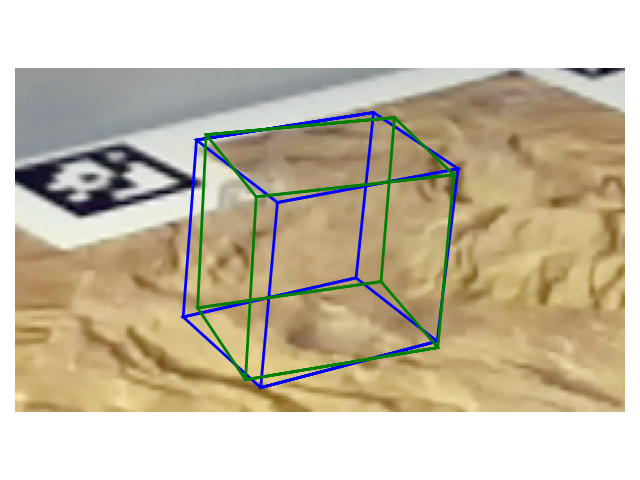}
  \end{subfigure}
  \caption{object mug$_1$ over texture 8}
  \label{fig:qr:8}
\end{subfigure}%
\hspace*{\fill}
\begin{subfigure}{.39\linewidth}
  \centering
  \begin{subfigure}{0.495\linewidth}
       \includegraphics[width=\linewidth,trim={0.5cm 2cm 0.5cm 2cm},clip]{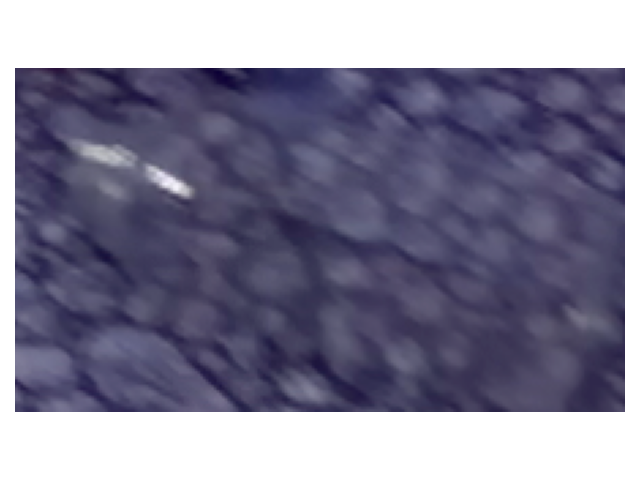}
  \end{subfigure}%
  \hspace*{\fill}
  \begin{subfigure}{0.495\linewidth}
      \includegraphics[width=\linewidth,trim={0.5cm 2cm 0.5cm 2cm},clip]{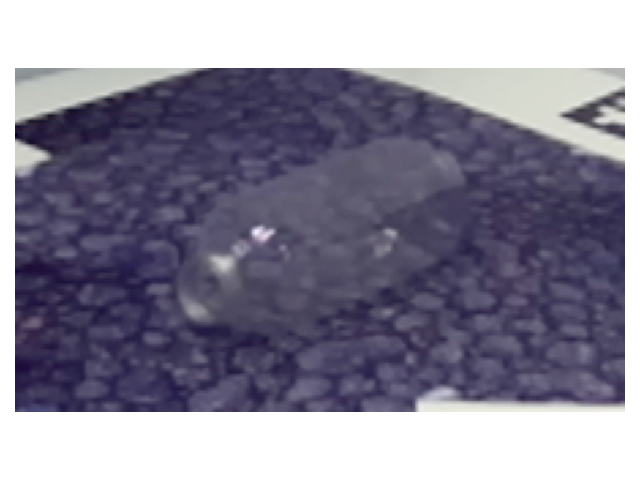}
  \end{subfigure}
  \par\smallskip
  \begin{subfigure}{0.495\linewidth}
      \includegraphics[width=\linewidth,trim={0.5cm 2cm 0.5cm 2cm},clip]{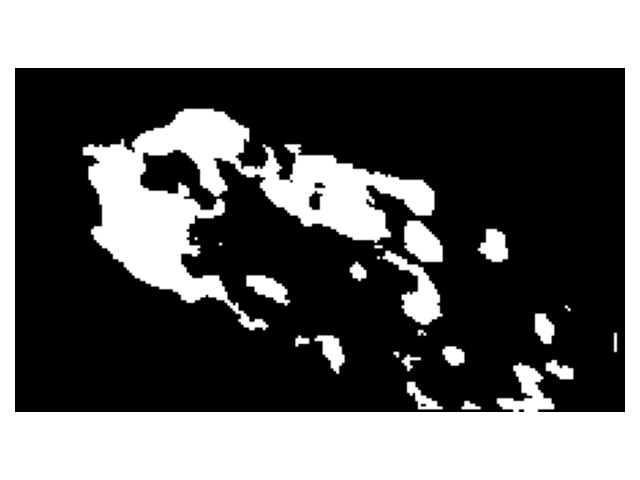}
  \end{subfigure}%
  \hspace*{\fill}
  \begin{subfigure}{0.495\linewidth}
      \includegraphics[width=\linewidth,trim={0.5cm 2cm 0.5cm 2cm},clip]{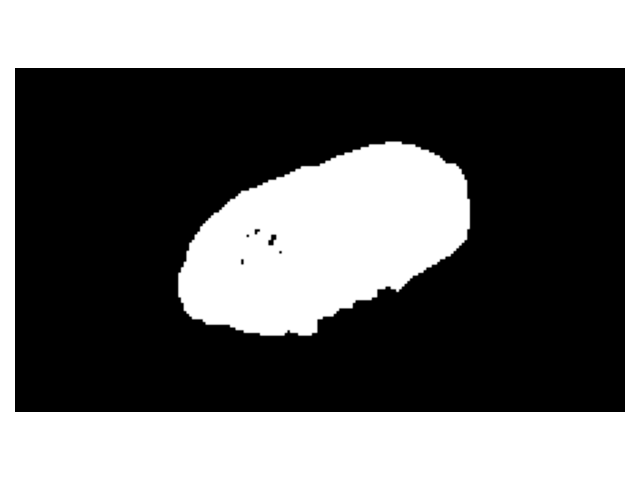}
  \end{subfigure}
  \par\smallskip
  \begin{subfigure}{0.495\linewidth}
      \includegraphics[width=\linewidth,trim={0.5cm 2cm 0.5cm 2cm},clip]{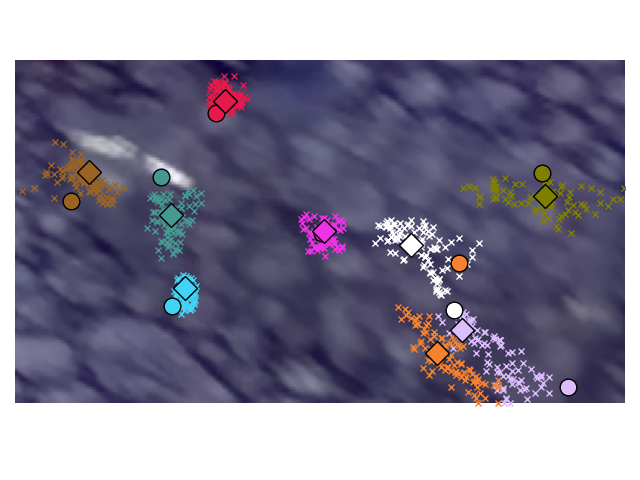}
  \end{subfigure}%
  \hspace*{\fill}
  \begin{subfigure}{0.495\linewidth}
      \includegraphics[width=\linewidth,trim={0.5cm 2cm 0.5cm 2cm},clip]{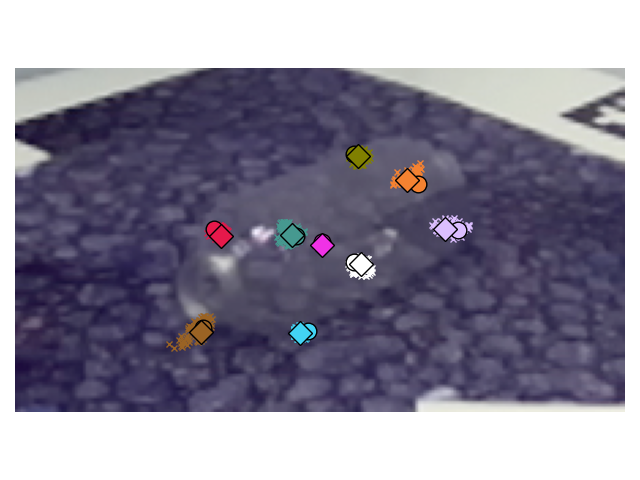}
  \end{subfigure}
  \par\smallskip
  \begin{subfigure}{0.495\linewidth}
      \includegraphics[width=\linewidth,trim={0.5cm 2cm 0.5cm 2cm},clip]{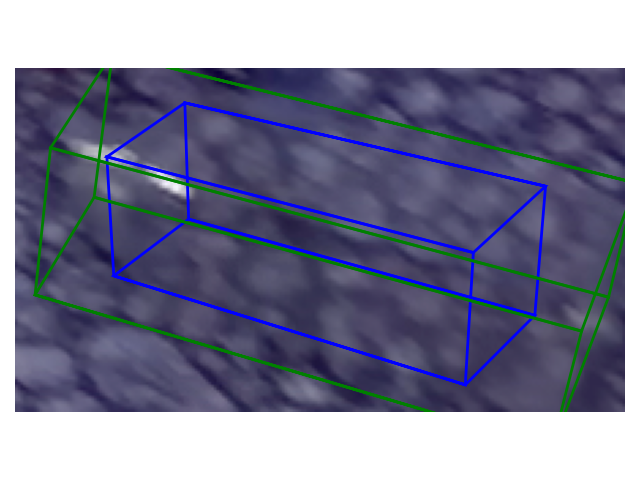}
  \end{subfigure}%
  \hspace*{\fill}
  \begin{subfigure}{0.495\linewidth}
      \includegraphics[width=\linewidth,trim={0.5cm 2cm 0.5cm 2cm},clip]{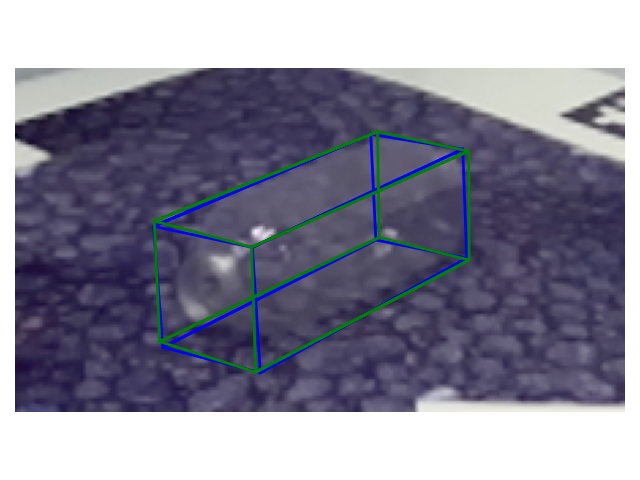}
  \end{subfigure}
  \caption{object bottle$_2$ over texture 4}
  \label{fig:qr:4}
\end{subfigure}%
\hspace*{\fill}
\begin{subfigure}{.195\linewidth}
  \centering
  \begin{subfigure}{0.99\linewidth}
       \includegraphics[width=\linewidth,trim={0.5cm 2cm 0.5cm 2cm},clip]{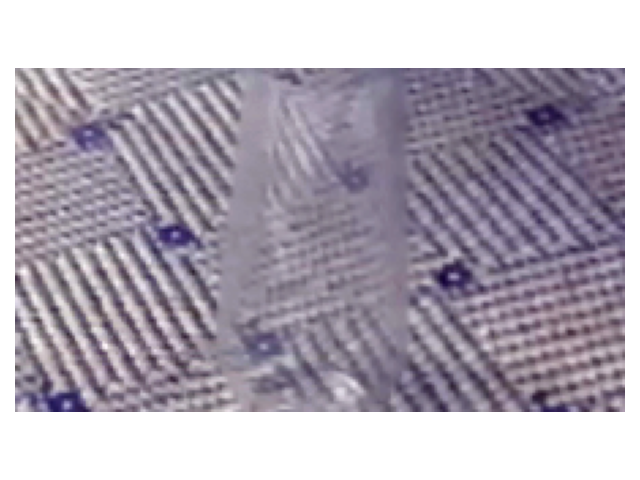}
  \end{subfigure}
  \par\smallskip
  \begin{subfigure}{0.99\linewidth}
       \includegraphics[width=\linewidth,trim={0.5cm 2cm 0.5cm 2cm},clip]{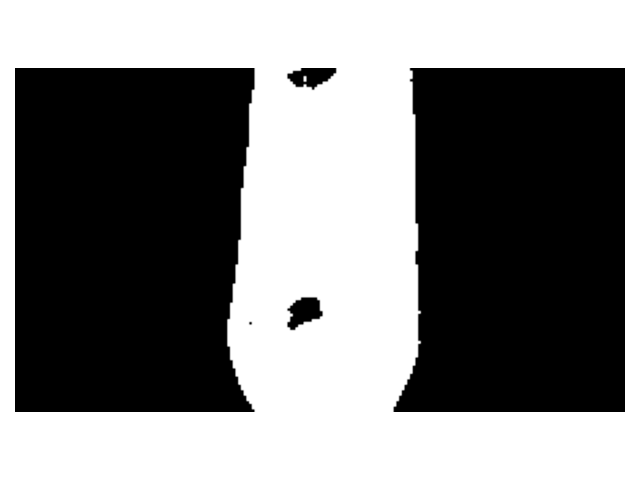}
  \end{subfigure}
  \par\smallskip
  \begin{subfigure}{0.99\linewidth}
       \includegraphics[width=\linewidth,trim={0.5cm 2cm 0.5cm 2cm},clip]{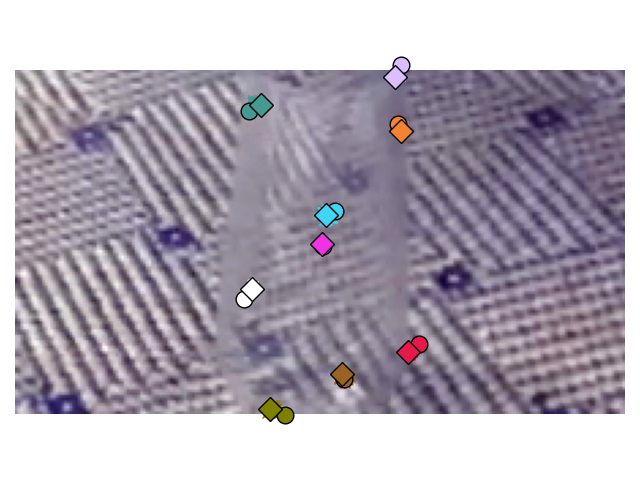}
  \end{subfigure}
  \par\smallskip
  \begin{subfigure}{0.99\linewidth}
       \includegraphics[width=\linewidth,trim={0.5cm 2cm 0.5cm 2cm},clip]{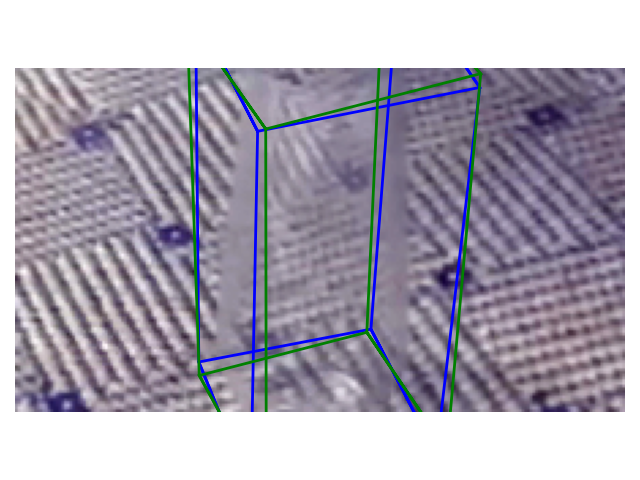}
  \end{subfigure}
  \caption{object cup$_1$ over texture 2}
  \label{fig:qr:2}
\end{subfigure}%

\caption{Qualitative results of our method. (First row) Left input image; (Second row) Predicted segmentation mask; (Third row) Predicted (diamond) vs ground truth (circle) keypoints + hypotheses variance; (Fourth row) Predicted (blue) vs ground truth (green) 3D bounding boxes. In \figref{fig:qr:8} and \figref{fig:qr:4},
we report an incorrect estimation in the left column and a correct one in the right column. \figref{fig:qr:8} displays the most common source of wrong pose estimation for our method, which is a bad object segmentation. In particular, texture 8 has proven to be the hardest in the dataset, due to its peculiar pattern. Similarly, in \figref{fig:qr:4} the object is badly segmented due to a high level of motion blur in the image, causing high variance in keypoints close to the bottom of the bottle and, consequently, an inaccurate object scale. Finally, \figref{fig:qr:2} shows how choosing a canonical pose allows the network to learn accurate keypoints positions for symmetric objects, with no rotation ambiguity.}
\label{fig:qual_res}
\vspace{-0.5cm}
\end{figure*}

We evaluated KVN on the TOD dataset and compared its performance with three other state-of-the-art object pose estimation methods: DenseFusion \cite{8953386}, which is based on monocular RGB-D images, KeyPose \cite{liu2020keypose} and GhostPose \cite{9636459}, both based on RGB stereo images. 
We also report the results obtained by training the standard PVNet without the differentiable RANSAC component, applied to stereo images with the UM-PnP procedure: in the tables we call this strategy s-PVNet. 
The evaluation results for KeyPose and DenseFusion are taken from \cite{liu2020keypose} and those for GhostPose from \cite{9636459}.
\tabref{tab:full-results} showcases the performance of all methods on the TOD dataset. GhostPose only focused on dishes (bottles, cups, and mugs) so there are no metrics for the objects ball$_0$, heart$_0$, and tree$_0$. The results show that our method performs comparably or better than the others on most objects, in particular those which have a more detailed shape, such as bottle$_2$ or the mugs. \figref{fig:teaser}(a-d) and \figref{fig:qual_res} show some qualitative results of our method. The lower results of all keypoints-based approaches (KVN, s-PVNet, and KeyPose) on the mug$_1$ object are due to the specific shape of this mug. If the handle of the mug is partially occluded, it may happen that during the training phase, the keypoints close to the squared base converge to an incorrect local minimum which corresponds to a 90-degree rotation around the vertical axis.
The KVN architecture overall improves the accuracy of s-PVNet, with an increase of 2\% in average AUC and 3.3\% in average $<$2cm, and a decrease of 2.1 mm in average MAE.
Finally, we compared KVN with s-PVNet in TTD, reporting the results in \tabref{tab:TTD_results}, where we replaced the AUC metric with ADD(-S) since the size of the dataset is small, while the MAE metric is omitted because the occlusion of some objects results in non-significant MAE values. 
Again KVN outperforms s-PVNet, especially with the most challenging objects ("coffee cup" and "candle holder", e.g., \figref{fig:teaser}(e,f)). s-PVNet surpasses KVN only on the simplest and largest object ("wine glass"). This can be partly explained by the fact that, while s-PVNet tries to converge all vectors independently to point to the same keypoint, where a "good" subset is then selected by the classical RANSAC step, KVN tries to converge all the keypoint hypotheses, even those generated from couple of points far from the keypoint, possibly less accurate than nearby ones. The presence of such hypotheses can sometimes negatively affect the accuracy of the estimate.
\subsubsection{Runtime}
At inference time, given that there is no need to back-propagate gradients, we perform keypoint prediction as in PVNet, so we do not increase the complexity of this step. Keypoint positions are predicted with a hypotheses pool of size 256 (128 in TTD), while keypoint covariance matrices are computed from a pool of size 1024. The average time required by our model for vote prediction+RANSAC+stereo PnP on a Titan RTX GPU and i7 CPU is 40 ms. KeyPose, which predicts 3D keypoint positions with a single network inference pass, requires only 3 ms for stereo pair on an NVidia Titan V GPU and an i7 desktop. GhostPose does not list inference time.

\section{Conclusion}

In this work, we introduced KVN, a novel stereo pose estimation pipeline based on robust keypoints correspondences. We first extended the original monocular PVNet architecture with a differentiable RANSAC layer, allowing the network to be trained directly on keypoint error. We then extended it to the stereo domain by fusing multiple robust monocular keypoint predictions with an uncertainty-driven multi-view PnP optimizer. We tested the effectiveness of our approach on the challenging task of transparent object pose estimation on two challenging datasets. Evaluation results show that our method achieves state-of-the-art accuracy.\\
Our method could be extended by integrating \cite{wei2022fully}'s minimal set sampling strategy for differentiable RANSAC, allowing dense gradients when back-propagating through RANSAC hypotheses. Additionally, a differentiable PnP layer, such as \cite{campbell2020solving,9156614} could be added after the differentiable RANSAC layer to train KVN end-to-end.
\bibliographystyle{IEEEtran}
\vspace{-0.1cm}
\bibliography{bibliography.bib}

\ifarXiv
        

\section*{Supplementary material}

# KVN: Keypoints Voting Network with Differentiable RANSAC for Stereo Pose Estimation (Supplementary Material)

Ivano Donadi and Alberto Pretto

*Abstract*—This document provides additional details about the experimental protocol, a description of the Transparent Tableware Dataset (TTD) dataset we created, further qualitative pose estimation results, and comprehensive per-object quantitative results on the public TOD dataset.

## I. Experimental Protocol Details

### A. Dataset Preprocessing

In the TOD dataset [1] used in the experiments, the ground truth pose annotation is stored as the 3D positions of a set of keypoints previously defined on the object's CAD model, while our method infers the 6-DoF object pose. For this reason, we re-label TOD by adding the ground truth transformation matrix obtained by solving an Orthogonal Procrustes problem between the CAD model's keypoints and the current image's keypoints. We also generate ground truth segmentation masks for the right camera images by projecting the object's mesh into the image plane by using the ground truth 6-DoF object pose.

### B. Image cropping

As in Keypose [1], our method assumes a detection stage that approximately determines the location of the object in the two images. This assumption is realized by extracting a Region Of Interest (ROI) around the object centroid of fixed size 126x224 pixels. All approaches compared below follow this protocol.

### C. Symmetric objects

In TOD, there are 5 objects with full 360° symmetry about their vertical axis: $ball_0$, $cup_0$, $cup_1$, $bottle_0$ and $bottle_1$. The pose of such objects can only be defined up to a rotation about their symmetry axis. To allow learning, as in [2], we define their canonical orientation to be the one in which the object x-axis is orthogonal to the camera z-axis (e.g., see Fig. 5b). We do not apply this procedure to the $bottle_2$ object since it has squared sides and so it loses full 360° symmetry.

## II. Transparent Tableware Dataset

To create the TTD dataset[1], we arranged 5 transparent tableware objects (a candle holder, a coffee cup, a glass, a little bottle, and a wine glass) on a set table together with

[1]Available at: https://doi.org/10.5281/zenodo.10580443

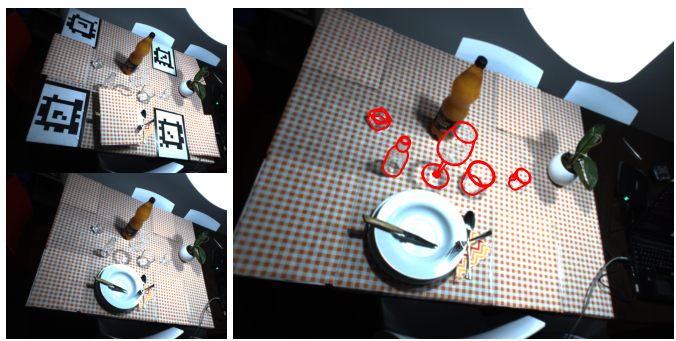

Fig. 1: Pose annotation strategy used in TTD: (Top left) Image with visible fiducial markers; (Bottom left) Dataset image acquired in the same position as the previous but with covered fiducial markers; (Right) Example of object pose annotations.

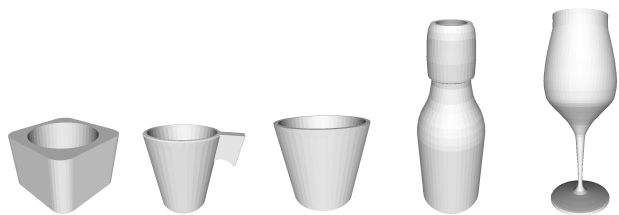

Fig. 2: The CAD models of the objects of TTD.

other householding objects (background objects). We set the table in four different ways, keeping (in different positions) all 5 transparent objects but changing the paper tablecloth and the other objects. For each of the four configurations, we collected 150 stereo images, randomly changing for each image the positions of the background objects. We moved a calibrated Flir Bumblebee 2 stereo camera mounted on a tripod to different positions around the table. To estimate the camera position, we attached four different AprilTag fiducial markers [3] at known positions on the table. The poses were obtained by using the PnP (Perspective-n-Point) algorithm. To avoid perceptual bias due to the presence of fiducial markers in the images, for each pose we collected two images: one with visible markers, and the other with the markers covered by pieces of tablecloth (see Fig. 1, left). We estimated the camera pose on the image with the markers, while we stored in the dataset the corresponding image without them. We provide the CAD models of the objects (see Fig. 2), the pose annotations (e.g., Fig. 1, right), and the segmentation masks.

| Object | scoring function | <2cm | MAE | AUC |
|---|---|---|---|---|
| heart$_0$ | $\sigma_1(*; \beta=1.7, \tau=0.99)$ | **87.3** | 10.0 | 90.5 |
| | $\sigma_2(*; \beta=100, \tau=0.99)$ | **87.3** | **9.8** | **90.7** |
| | $\sigma_3(*; \beta=6, \tau=0.5)$ | 86.8 | 9.9 | 90.5 |
| | $f(*; t=0.9, v=0.1)$ | 87.2 | 10.0 | 90.4 |
| mug$_6$ | $\sigma_1(*; \beta=1.7, \tau=0.99)$ | 89.1 | 14.5 | 86.9 |
| | $\sigma_2(*; \beta=100, \tau=0.99)$ | 88.5 | 14.2 | 87.0 |
| | $\sigma_3(*; \beta=6, \tau=0.5)$ | 85.8 | 17.6 | 84.5 |
| | $f(*; t=0.9, v=0.1)$ | **90.7** | **13.2** | **88.3** |
| cup$_1$ | $\sigma_1(*; \beta=1.7, \tau=0.99)$ | 97.7 | 5.4 | 95.1 |
| | $\sigma_2(*; \beta=100, \tau=0.99)$ | 97.5 | 5.7 | 94.7 |
| | $\sigma_3(*; \beta=6, \tau=0.5)$ | **98.0** | **5.0** | **95.4** |
| | $f(*; t=0.9, v=0.1)$ | 97.3 | 6.3 | 94.2 |
| Avg. | $\sigma_1(*; \beta=1.7, \tau=0.99)$ | 91.3 | 9.9 | 90.8 |
| | $\sigma_2(*; \beta=100, \tau=0.99)$ | 91.1 | 9.9 | 90.8 |
| | $\sigma_3(*; \beta=6, \tau=0.5)$ | 90.2 | 10.8 | 90.1 |
| | $f(*; t=0.9, v=0.1)$ | **91.7** | **9.8** | **90.9** |

TABLE I: Evaluation of different scoring functions for differentiable RANSAC.

## III. SCORING FUNCTION EVALUATION

We report in Tab. I the average *per object* values of MAE, AUC, and <2cm metrics for the four different scoring functions we tested on a representative for each of the three object categories present in the dataset: mirror symmetrical (heart$_0$), asymmetrical (mug$_6$) and symmetrical w.r.t. an axis (cup$_1$).

## IV. QUALITATIVE RESULTS

Fig. 3 shows a qualitative comparison of the quality of predictions between PVNet and KVN. In Fig. 4 we present some positive KVN evaluation results on TOD objects, while Fig. 5 shows some of KVN failure cases.

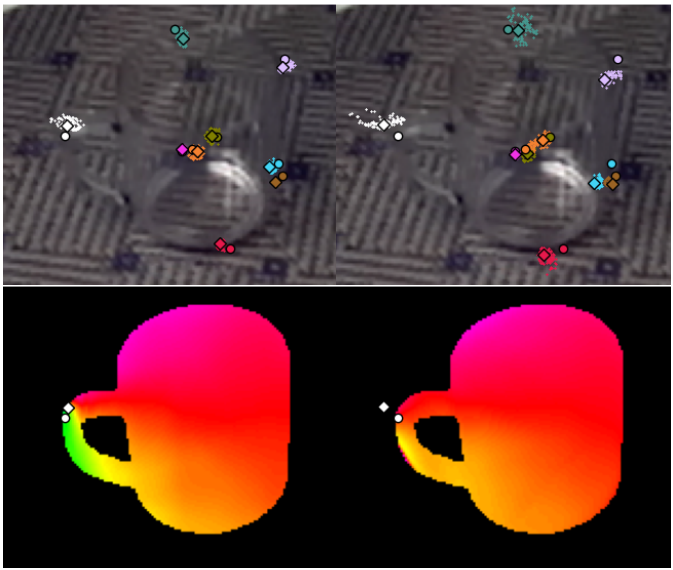

Fig. 3: The precision and variance of RANSAC hypotheses for KVN and PVNet are compared in the left and right columns, respectively. The top row displays the top 100 RANSAC hypotheses for all keypoints, while the bottom row shows the predicted mug handle keypoint vote direction for each pixel as hue values. The ground truth keypoints are depicted as circles, while the predicted ones are represented as diamonds. KVN presents lower keypoint variance in general and is capable of preserving the correct geometrical relationship between the orange and green keypoints and between the brown and cyan keypoints, which is not retained in PVNet.

## V. DETAILED QUANTITATIVE RESULTS

In this section, we report the per-texture results on all TOD objects. Tab. II to Tab. XVI contain the results for the PVNet+UM-PnP approach, while Tab. XVII to Tab. XXXI contain the results for the PVNet+UM-PnP method. Each row, labeled T$i$, contains the average metrics for ten consecutive evaluations of the model trained on all textures but texture $i$, which is used for testing instead. We provide the following metrics:

- Projection 2D (P2D): the percentage of test samples for which the average distance between ground truth and estimated 2D keypoint reprojections is smaller than a given threshold, in our case 5 pixels;
- ADD(-S): the percentage of test samples for which the average distance between ground truth and estimated 3D keypoint positions is smaller than a percentage of the object's diameter, in our case 10%. If the object is symmetric, the ambiguity in the keypoints association is solved by pairing each estimated keypoint with its closest ground truth counterpart;
- <2cm: as ADD(-S) but the threshold is fixed at 2cm;
- Mean Absolute Error (MAE): the average absolute 3D position error between ground truth and estimated keypoints (in mm);
- 5 centimeters 5 degrees (5c5d): the percentage of samples for which the rotation error is lower than 5 degrees and the translation error is lower than 5cm, computed on the transformation matrix;
- Area Under Curve (AUC): area under the curve in which each point represents the ADD(-S) metric obtained using its x-axis value as the threshold: in our case the x-axis ranges from 0cm to 10cm;
- MaskAP70 (AP70): the percentage of samples for which at least 70% of the pixels in the segmentation mask are labeled correctly.

All metrics are expressed as percentages except for the MAE, which is expressed in millimeters. When computing the aforementioned metrics, we used the set of 3D keypoints defined in the TOD dataset for each object.

| ball$_0$ | P2D | ADD(-S) | <2cm | MAE | 5c5d | AUC | AP70 |
|---|---|---|---|---|---|---|---|
| T0 | 98.7 | 97.8 | 98.1 | 3.4 | 33.9 | 97.1 | 99.7 |
| T1 | 99.2 | 97.5 | 98.7 | 3.5 | 34.8 | 97.0 | 99.7 |
| T2 | 97.4 | 95.6 | 96.6 | 4.2 | 41.5 | 96.2 | 1.00 |
| T3 | 98.1 | 95.7 | 96.1 | 4.3 | 28.1 | 96.2 | 99.4 |
| T4 | 98.1 | 94.9 | 97.3 | 4.4 | 44.3 | 96.0 | 1.00 |
| T5 | 99.6 | 92.6 | 98.2 | 4.5 | 34.8 | 96.0 | 1.00 |
| T6 | 98.4 | 96.5 | 97.5 | 3.9 | 43.0 | 96.5 | 1.00 |
| T7 | 99.2 | 93.5 | 97.2 | 4.1 | 45.2 | 96.4 | 1.00 |
| T8 | 98.2 | 96.8 | 98.5 | 3.6 | 62.2 | 96.9 | 1.00 |
| T9 | 99.7 | 97.4 | 98.4 | 3.2 | 18.4 | 97.3 | 1.00 |
| Avg. | 98.7 | 95.8 | 97.7 | 3.9 | 38.6 | 96.6 | 99.9 |

TABLE II: PVNet+UM-PnP results on ball$_0$ object

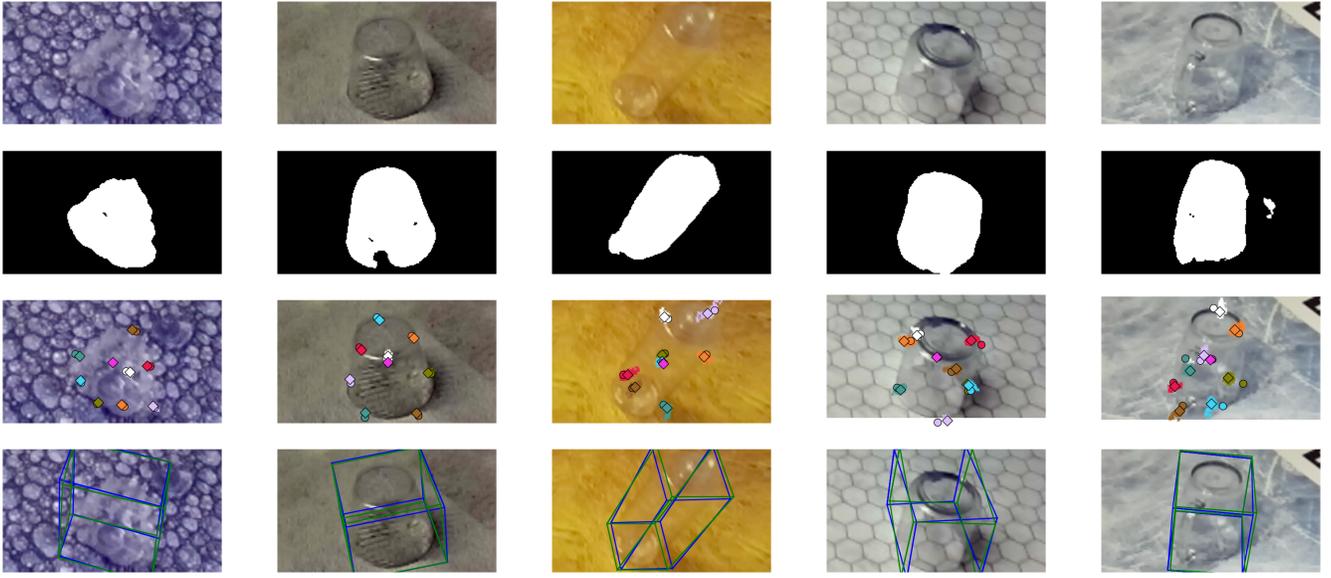

(a) object tree$_0$ over texture 4  (b) object cup$_0$ over texture 1  (c) object bottle$_1$ over texture 3  (d) object mug$_2$ over texture 6  (e) object mug$_3$ over texture 7

Fig. 4: Successful KVN pose estimations. (First row) Left input image; (Second row) Predicted segmentation mask; (Third row) Predicted (diamond) vs ground truth (circle) keypoints + hypotheses variance; (Fourth row) Predicted (blue) vs ground truth (green) 3D bounding boxes.

| bottle$_0$ | P2D | ADD(-S) | <2cm | MAE | 5c5d | AUC | AP70 |
|---|---|---|---|---|---|---|---|
| T0 | 98.2 | 96.7 | 98.1 | 4.4 | 80.2 | 96.1 | 1.00 |
| T1 | 94.7 | 93.9 | 94.2 | 5.7 | 53.6 | 94.7 | 1.00 |
| T2 | 97.9 | 97.3 | 97.6 | 4.6 | 62.2 | 95.9 | 1.00 |
| T3 | 97.8 | 97.0 | 97.6 | 4.4 | 70.3 | 96.0 | 1.00 |
| T4 | 95.2 | 93.6 | 96.0 | 5.9 | 56.0 | 94.6 | 1.00 |
| T5 | 95.4 | 93.4 | 94.9 | 6.1 | 59.1 | 94.3 | 1.00 |
| T6 | 92.7 | 92.4 | 92.8 | 6.6 | 51.7 | 93.9 | 1.00 |
| T7 | 93.3 | 90.5 | 93.0 | 7.3 | 49.7 | 93.1 | 1.00 |
| T8 | 93.8 | 90.3 | 94.3 | 7.0 | 39.0 | 93.4 | 99.7 |
| T9 | 93.5 | 93.5 | 93.5 | 5.7 | 59.6 | 94.8 | 1.00 |
| Avg. | 95.2 | 93.9 | 95.2 | 5.8 | 58.1 | 94.7 | 1.00 |

TABLE III: PVNet+UM-PnP results on bottle_0 object

| bottle$_2$ | P2D | ADD(-S) | <2cm | MAE | 5c5d | AUC | AP70 |
|---|---|---|---|---|---|---|---|
| T0 | 90.6 | 93.6 | 93.8 | 8.8 | 63.6 | 91.6 | 91.2 |
| T1 | 97.9 | 96.7 | 96.8 | 6.9 | 76.7 | 93.6 | 1.00 |
| T2 | 87.2 | 88.5 | 89.5 | 9.7 | 56.6 | 90.8 | 88.0 |
| T3 | 62.0 | 83.1 | 83.9 | 16.4 | 40.1 | 84.3 | 95.5 |
| T4 | 84.4 | 93.9 | 94.1 | 9.4 | 57.0 | 91.1 | 88.0 |
| T5 | 97.0 | 98.4 | 98.4 | 6.2 | 81.6 | 94.3 | 97.2 |
| T6 | 83.7 | 92.3 | 93.0 | 9.4 | 53.0 | 91.1 | 91.2 |
| T7 | 84.0 | 93.2 | 93.7 | 8.8 | 66.1 | 91.6 | 93.8 |
| T8 | 60.1 | 76.4 | 77.3 | 19.0 | 47.1 | 82.3 | 88.7 |
| T9 | 86.9 | 90.0 | 90.1 | 12.2 | 48.8 | 88.3 | 96.7 |
| Avg. | 83.4 | 90.6 | 91.1 | 10.7 | 59.1 | 89.9 | 93.0 |

TABLE V: PVNet+UM-PnP results on bottle_2 object

| bottle$_1$ | P2D | ADD(-S) | <2cm | MAE | 5c5d | AUC | AP70 |
|---|---|---|---|---|---|---|---|
| T0 | 96.8 | 98.1 | 98.1 | 5.8 | 90.6 | 94.7 | 1.00 |
| T1 | 93.9 | 97.3 | 97.3 | 5.8 | 84.7 | 94.7 | 1.00 |
| T2 | 88.2 | 94.5 | 94.6 | 7.9 | 82.9 | 92.6 | 1.00 |
| T3 | 89.0 | 96.7 | 96.9 | 7.2 | 83.2 | 93.2 | 1.00 |
| T4 | 84.8 | 93.9 | 94.3 | 8.1 | 73.7 | 92.3 | 99.4 |
| T5 | 94.1 | 97.5 | 97.5 | 5.9 | 88.6 | 94.6 | 1.00 |
| T6 | 93.3 | 97.2 | 97.2 | 6.4 | 86.4 | 94.1 | 1.00 |
| T7 | 95.0 | 97.4 | 97.5 | 5.9 | 87.9 | 94.6 | 1.00 |
| T8 | 87.5 | 97.0 | 97.2 | 6.4 | 78.2 | 94.0 | 1.00 |
| T9 | 84.9 | 95.1 | 95.2 | 8.0 | 76.6 | 92.6 | 1.00 |
| Avg. | 90.7 | 96.5 | 96.6 | 6.7 | 83.3 | 93.7 | 99.9 |

TABLE IV: PVNet+UM-PnP results on bottle_1 object

| cup$_0$ | P2D | ADD(-S) | <2cm | MAE | 5c5d | AUC | AP70 |
|---|---|---|---|---|---|---|---|
| T0 | 97.0 | 95.9 | 96.8 | 6.6 | 32.9 | 93.9 | 1.00 |
| T1 | 97.8 | 94.3 | 96.7 | 7.3 | 25.5 | 93.2 | 1.00 |
| T2 | 95.7 | 92.9 | 97.7 | 8.2 | 27.6 | 92.4 | 97.8 |
| T3 | 99.5 | 99.0 | 99.5 | 5.1 | 28.3 | 95.4 | 1.00 |
| T4 | 97.4 | 97.9 | 98.4 | 6.3 | 37.0 | 94.3 | 1.00 |
| T5 | 96.8 | 94.7 | 96.5 | 7.8 | 44.7 | 92.8 | 1.00 |
| T6 | 96.3 | 94.8 | 97.2 | 7.6 | 27.0 | 92.9 | 1.00 |
| T7 | 94.0 | 89.3 | 93.6 | 9.2 | 28.6 | 91.3 | 96.3 |
| T8 | 86.0 | 88.5 | 93.2 | 9.9 | 11.1 | 90.7 | 91.2 |
| T9 | 95.0 | 95.6 | 98.5 | 6.7 | 24.5 | 93.8 | 1.00 |
| Avg. | 94.7 | 94.3 | 96.8 | 7.5 | 28.7 | 93.1 | 98.5 |

TABLE VI: PVNet+UM-PnP results on cup_0 object

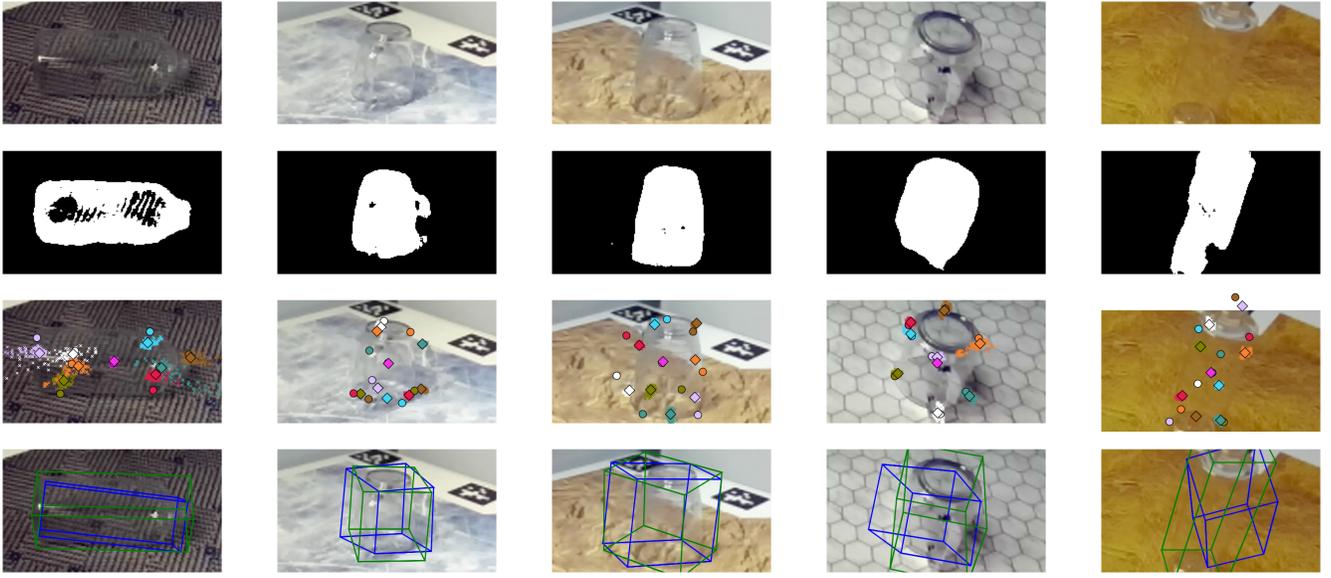

(a) object bottle$_2$ over texture 2  (b) object mug$_3$ over texture 7  (c) object mug$_4$ over texture 8  (d) object mug$_2$ over texture 6  (e) object bottle$_1$ over texture 3

Fig. 5: Unsuccessful KVN pose estimations. (First row) Left input image; (Second row) Predicted segmentation mask; (Third row) Predicted (diamond) vs ground truth (circle) keypoints + hypotheses variance; (Fourth row) Predicted (blue) vs ground truth (green) 3D bounding boxes. Most issues can be explained by the failure of the model to detect crucial attributes, such as the handle for mugs or the neck for bottles. As a result, this introduces ambiguity in determining the object's pose.

| cup$_1$ | P2D | ADD(-S) | <2cm | MAE | 5c5d | AUC | AP70 |
|---|---|---|---|---|---|---|---|
| T0 | 98.8 | 98.8 | 98.8 | 4.5 | 56.1 | 96.0 | 1.00 |
| T1 | 98.8 | 98.8 | 98.8 | 4.1 | 87.1 | 96.3 | 1.00 |
| T2 | 95.1 | 98.0 | 98.6 | 5.9 | 56.7 | 94.6 | 1.00 |
| T3 | 97.8 | 98.9 | 98.9 | 4.8 | 83.5 | 95.6 | 1.00 |
| T4 | 79.5 | 92.7 | 93.7 | 8.8 | 38.8 | 91.6 | 96.3 |
| T5 | 98.0 | 98.6 | 98.6 | 5.0 | 77.4 | 95.4 | 1.00 |
| T6 | 97.9 | 98.4 | 98.4 | 5.0 | 77.1 | 95.4 | 1.00 |
| T7 | 87.6 | 89.4 | 91.1 | 9.9 | 53.2 | 90.5 | 98.1 |
| T8 | 96.9 | 96.3 | 96.8 | 6.8 | 46.3 | 93.6 | 1.00 |
| T9 | 96.3 | 96.3 | 96.9 | 6.5 | 54.7 | 93.9 | 1.00 |
| Avg. | 94.7 | 96.6 | 97.1 | 6.1 | 63.1 | 94.3 | 99.4 |

TABLE VII: PVNet+UM-PnP results on cup$_1$ object

| mug$_1$ | P2D | ADD(-S) | <2cm | MAE | 5c5d | AUC | AP70 |
|---|---|---|---|---|---|---|---|
| T0 | 62.8 | 65.8 | 71.9 | 21.7 | 16.8 | 79.0 | 1.00 |
| T1 | 47.8 | 56.7 | 65.5 | 22.1 | 18.0 | 78.3 | 1.00 |
| T2 | 50.3 | 52.9 | 69.4 | 23.0 | 0.7 | 78.8 | 1.00 |
| T3 | 53.9 | 72.1 | 84.7 | 15.5 | 25.0 | 85.2 | 1.00 |
| T4 | 66.4 | 68.8 | 77.4 | 17.0 | 17.4 | 83.5 | 98.8 |
| T5 | 76.3 | 77.9 | 84.8 | 14.5 | 31.0 | 86.2 | 1.00 |
| T6 | 46.9 | 49.0 | 64.7 | 25.4 | 12.3 | 76.7 | 97.9 |
| T7 | 64.8 | 78.5 | 87.0 | 16.4 | 1.8 | 84.4 | 1.00 |
| T8 | 18.5 | 26.3 | 39.7 | 43.5 | 2.8 | 62.8 | 87.9 |
| T9 | 75.0 | 65.4 | 73.8 | 18.3 | 28.1 | 82.2 | 1.00 |
| Avg. | 56.3 | 61.3 | 71.9 | 21.7 | 15.4 | 79.7 | 98.5 |

TABLE IX: PVNet+UM-PnP results on mug$_1$ object

| mug$_0$ | P2D | ADD(-S) | <2cm | MAE | 5c5d | AUC | AP70 |
|---|---|---|---|---|---|---|---|
| T0 | 87.6 | 92.0 | 94.5 | 10.4 | 29.3 | 90.7 | 1.00 |
| T1 | 76.2 | 75.1 | 82.0 | 15.1 | 15.9 | 85.9 | 1.00 |
| T2 | 79.6 | 84.2 | 89.3 | 14.1 | 39.4 | 87.1 | 1.00 |
| T3 | 22.2 | 42.9 | 52.5 | 36.2 | 2.6 | 70.2 | 79.7 |
| T4 | 72.7 | 71.9 | 80.1 | 17.8 | 19.1 | 83.0 | 1.00 |
| T5 | 86.2 | 93.4 | 95.6 | 9.3 | 46.6 | 91.8 | 1.00 |
| T6 | 88.9 | 93.8 | 97.0 | 10.0 | 43.4 | 90.6 | 1.00 |
| T7 | 30.8 | 59.0 | 79.3 | 16.2 | 4.2 | 84.5 | 99.4 |
| T8 | 69.2 | 62.4 | 78.1 | 19.2 | 34.9 | 82.7 | 1.00 |
| T9 | 94.9 | 94.4 | 96.1 | 10.2 | 49.2 | 91.8 | 1.00 |
| Avg. | 70.8 | 76.9 | 84.4 | 15.8 | 28.4 | 85.8 | 97.9 |

TABLE VIII: PVNet+UM-PnP results on mug$_0$ object

| mug$_2$ | P2D | ADD(-S) | <2cm | MAE | 5c5d | AUC | AP70 |
|---|---|---|---|---|---|---|---|
| T0 | 95.6 | 95.4 | 96.2 | 10.3 | 808 | 91.2 | 1.00 |
| T1 | 97.0 | 95.6 | 97.6 | 8.7 | 669 | 92.3 | 1.00 |
| T2 | 62.3 | 70.3 | 78.6 | 16.5 | 29.2 | 84.8 | 95.9 |
| T3 | 55.9 | 64.8 | 71.4 | 20.8 | 17.2 | 80.5 | 96.9 |
| T4 | 67.2 | 78.0 | 82.6 | 14.8 | 13.2 | 86.1 | 96.3 |
| T5 | 92.2 | 91.2 | 93.8 | 10.4 | 52.4 | 90.6 | 97.2 |
| T6 | 85.9 | 82.5 | 86.8 | 13.6 | 37.8 | 87.6 | 1.00 |
| T7 | 80.4 | 82.5 | 90.1 | 12.9 | 29.2 | 88.2 | 1.00 |
| T8 | 96.0 | 91.2 | 95.5 | 10.7 | 59.5 | 91.0 | 95.4 |
| T9 | 95.7 | 91.7 | 94.2 | 9.1 | 61.1 | 91.8 | 98.1 |
| Avg. | 82.8 | 84.3 | 88.7 | 12.8 | 44.7 | 88.4 | 98.0 |

TABLE X: PVNet+UM-PnP results on mug$_2$ object

| $mug_3$ | P2D | ADD(-S) | <2cm | MAE | 5c5d | AUC | AP70 |
|---|---|---|---|---|---|---|---|
| T0 | 90.3 | 92.1 | 92.4 | 14.3 | 64.7 | 88.6 | 91.2 |
| T1 | 93.3 | 93.1 | 93.8 | 11.6 | 51.0 | 90.5 | 98.8 |
| T2 | 62.0 | 69.1 | 70.6 | 24.7 | 37.6 | 77.7 | 96.3 |
| T3 | 94.7 | 95.6 | 95.9 | 10.1 | 81.2 | 91.8 | 98.8 |
| T4 | 72.9 | 84.8 | 86.7 | 15.9 | 36.0 | 85.7 | 72.5 |
| T5 | 84.4 | 86.4 | 88.6 | 14.2 | 43.5 | 87.7 | 80.3 |
| T6 | 93.5 | 96.0 | 96.2 | 8.8 | 82.1 | 92.5 | 99.7 |
| T7 | 74.7 | 82.0 | 83.5 | 19.5 | 51.2 | 83.8 | 68.4 |
| T8 | 57.6 | 77.8 | 80.3 | 17.7 | 20.0 | 83.3 | 95.3 |
| T9 | 78.1 | 95.1 | 96.6 | 10.7 | 53.1 | 91.2 | 94.7 |
| Avg. | 80.2 | 87.2 | 88.5 | 14.7 | 52.0 | 87.3 | 89.6 |

TABLE XI: PVNet+UM-PnP results on $mug_3$ object

| $heart_0$ | P2D | ADD(-S) | <2cm | MAE | 5c5d | AUC | AP70 |
|---|---|---|---|---|---|---|---|
| T0 | 95.9 | 91.3 | 96.2 | 6.4 | 36.9 | 94.1 | 1.00 |
| T1 | 90.9 | 87.6 | 91.9 | 8.1 | 18.5 | 92.3 | 1.00 |
| T2 | 72.5 | 64.6 | 78.3 | 14.5 | 12.7 | 86.1 | 99.4 |
| T3 | 90.7 | 86.0 | 91.4 | 8.9 | 27.0 | 91.5 | 1.00 |
| T4 | 83.8 | 73.9 | 88.2 | 10.4 | 13.6 | 90.0 | 1.00 |
| T5 | 64.7 | 62.3 | 66.8 | 16.5 | 34.8 | 83.9 | 1.00 |
| T6 | 84.6 | 83.3 | 94.6 | 7.9 | 11.4 | 92.5 | 1.00 |
| T7 | 95.6 | 94.2 | 95.6 | 6.2 | 46.0 | 94.2 | 1.00 |
| T8 | 73.3 | 67.9 | 86.5 | 11.1 | 9.3 | 89.3 | 1.00 |
| T9 | 56.4 | 53.4 | 65.0 | 19.4 | 13.0 | 80.9 | 1.00 |
| Avg. | 80.9 | 76.5 | 85.5 | 10.9 | 22.3 | 89.5 | 99.9 |

TABLE XV: PVNet+UM-PnP results on $heart_0$ object

| $mug_4$ | P2D | ADD(-S) | <2cm | MAE | 5c5d | AUC | AP70 |
|---|---|---|---|---|---|---|---|
| T0 | 86.3 | 96.3 | 96.1 | 10.0 | 54.6 | 90.8 | 99.1 |
| T1 | 89.4 | 98.2 | 98.2 | 7.4 | 62.7 | 93.2 | 99.2 |
| T2 | 72.9 | 93.7 | 93.3 | 11.4 | 30.9 | 89.7 | 74.7 |
| T3 | 69.2 | 95.0 | 94.4 | 11.2 | 25.3 | 89.6 | 74.4 |
| T4 | 85.3 | 89.5 | 89.2 | 13.8 | 55.6 | 87.8 | 38.8 |
| T5 | 84.4 | 97.9 | 97.6 | 8.5 | 47.1 | 92.3 | 97.2 |
| T6 | 66.1 | 80.3 | 79.3 | 16.0 | 44.3 | 85.0 | 92.8 |
| T7 | 84.6 | 94.9 | 94.3 | 10.0 | 60.7 | 90.7 | 28.7 |
| T8 | 75.9 | 89.7 | 88.9 | 13.5 | 21.2 | 87.9 | 1.00 |
| T9 | 58.4 | 87.8 | 87.2 | 14.1 | 34.2 | 86.6 | 99.1 |
| Avg. | 77.3 | 92.3 | 91.8 | 11.6 | 43.7 | 89.4 | 80.4 |

TABLE XII: PVNet+UM-PnP results on $mug_4$ object

| $tree_0$ | P2D | ADD(-S) | <2cm | MAE | 5c5d | AUC | AP70 |
|---|---|---|---|---|---|---|---|
| T0 | 82.1 | 92.8 | 93.4 | 10.0 | 61.5 | 91.6 | 1.00 |
| T1 | 90.3 | 97.1 | 99.1 | 7.0 | 20.0 | 93.4 | 1.00 |
| T2 | 80.6 | 90.6 | 93.0 | 9.9 | 33.6 | 90.5 | 1.00 |
| T3 | 83.2 | 98.3 | 99.6 | 6.4 | 31.8 | 94.0 | 1.00 |
| T4 | 96.8 | 97.0 | 98.1 | 5.6 | 51.3 | 94.9 | 1.00 |
| T5 | 71.9 | 82.4 | 83.0 | 13.4 | 51.2 | 87.1 | 1.00 |
| T6 | 59.7 | 72.7 | 78.0 | 19.8 | 19.5 | 80.7 | 99.6 |
| T7 | 86.8 | 91.4 | 96.5 | 8.7 | 56.4 | 91.7 | 1.00 |
| T8 | 54.7 | 80.1 | 88.0 | 15.6 | 19.2 | 85.9 | 98.7 |
| T9 | 94.5 | 98.9 | 98.9 | 4.9 | 60.2 | 95.6 | 1.00 |
| Avg. | 80.1 | 90.1 | 92.8 | 10.1 | 40.5 | 90.5 | 99.8 |

TABLE XVI: PVNet+UM-PnP results on $tree_0$ object

| $mug_5$ | P2D | ADD(-S) | <2cm | MAE | 5c5d | AUC | AP70 |
|---|---|---|---|---|---|---|---|
| T0 | 83.1 | 91.6 | 91.3 | 0146 | 65.9 | 87.6 | 95.3 |
| T1 | 82.8 | 88.9 | 88.7 | 19.7 | 67.8 | 82.7 | 1.00 |
| T2 | 83.1 | 91.6 | 91.6 | 15.4 | 69.3 | 87.4 | 72.9 |
| T3 | 74.1 | 90.3 | 90.1 | 16.1 | 50.2 | 86.1 | 76.2 |
| T4 | 80.8 | 90.4 | 89.9 | 14.9 | 51.8 | 86.9 | 54.9 |
| T5 | 89.2 | 93.2 | 93.0 | 13.0 | 76.8 | 88.4 | 40.3 |
| T6 | 73.6 | 90.6 | 90.2 | 16.0 | 47.3 | 86.3 | 92.8 |
| T7 | 77.3 | 81.6 | 81.4 | 25.7 | 44.7 | 78.7 | 27.2 |
| T8 | 83.4 | 93.0 | 92.7 | 13.6 | 53.1 | 88.7 | 70.3 |
| T9 | 87.2 | 93.8 | 93.8 | 13.9 | 53.3 | 87.3 | 97.5 |
| Avg. | 81.5 | 90.5 | 90.3 | 16.3 | 58.0 | 86.0 | 72.7 |

TABLE XIII: PVNet+UM-PnP results on $mug_5$ object

| $ball_0$ | P2D | ADD(-S) | <2cm | MAE | 5c5d | AUC | AP70 |
|---|---|---|---|---|---|---|---|
| T0 | 99.9 | 98.3 | 99.1 | 3.1 | 58.8 | 97.4 | 1.00 |
| T1 | 99.5 | 98.3 | 99.7 | 3.1 | 17.9 | 97.4 | 1.00 |
| T2 | 99.9 | 98.8 | 99.5 | 3.1 | 52.7 | 97.4 | 1.00 |
| T3 | 99.4 | 98.0 | 99.4 | 2.8 | 56.1 | 97.7 | 1.00 |
| T4 | 99.9 | 99.5 | 99.5 | 2.7 | 48.4 | 97.8 | 1.00 |
| T5 | 99.2 | 97.5 | 99.8 | 3.4 | 42.0 | 97.1 | 1.00 |
| T6 | 99.8 | 95.8 | 99.3 | 4.0 | 35.3 | 96.5 | 96.6 |
| T7 | 99.9 | 98.8 | 99.6 | 2.8 | 54.5 | 97.7 | 99.7 |
| T8 | 1.00 | 99.6 | 99.7 | 3.0 | 60.2 | 97.5 | 1.00 |
| T9 | 99.9 | 98.4 | 99.7 | 2.8 | 58.3 | 97.7 | 1.00 |
| Avg. | 99.7 | 98.3 | 99.5 | 3.1 | 48.4 | 97.4 | 99.6 |

TABLE XVII: KVN+UM-PnP results on $ball_0$ object

| $mug_6$ | P2D | ADD(-S) | <2cm | MAE | 5c5d | AUC | AP70 |
|---|---|---|---|---|---|---|---|
| T0 | 79.5 | 92.0 | 91.6 | 14.4 | 50.1 | 87.1 | 65.8 |
| T1 | 80.6 | 96.0 | 95.2 | 10.4 | 68.8 | 90.6 | 99.4 |
| T2 | 81.3 | 93.2 | 92.2 | 11.1 | 48.3 | 90.0 | 56.2 |
| T3 | 58.6 | 85.5 | 84.0 | 19.0 | 35.0 | 83.0 | 94.7 |
| T4 | 80.9 | 91.8 | 90.8 | 13.4 | 49.4 | 88.0 | 50.6 |
| T5 | 87.7 | 94.3 | 93.7 | 11.7 | 51.1 | 89.7 | 91.9 |
| T6 | 70.0 | 93.9 | 92.0 | 12.2 | 40.4 | 88.5 | 94.7 |
| T7 | 69.0 | 87.4 | 86.3 | 16.6 | 29.7 | 84.6 | 23.0 |
| T8 | 54.3 | 85.2 | 81.7 | 17.1 | 23.6 | 84.7 | 50.0 |
| T9 | 70.7 | 88.6 | 87.1 | 15.1 | 45.5 | 86.1 | 65.0 |
| Avg. | 73.3 | 90.8 | 89.5 | 14.1 | 44.2 | 87.2 | 69.1 |

TABLE XIV: PVNet+UM-PnP results on $mug_6$ object

| $bottle_0$ | P2D | ADD(-S) | <2cm | MAE | 5c5d | AUC | AP70 |
|---|---|---|---|---|---|---|---|
| T0 | 96.7 | 96.6 | 96.7 | 4.4 | 86.2 | 96.1 | 1.00 |
| T1 | 97.9 | 97.4 | 97.7 | 3.7 | 91.5 | 96.8 | 1.00 |
| T2 | 96.8 | 95.5 | 95.8 | 5.2 | 79.8 | 95.3 | 1.00 |
| T3 | 95.8 | 93.5 | 95.7 | 6.3 | 53.5 | 94.1 | 95.9 |
| T4 | 92.8 | 94.3 | 94.3 | 6.1 | 66.6 | 94.4 | 99.1 |
| T5 | 93.2 | 93.2 | 93.2 | 5.8 | 77.4 | 94.7 | 1.00 |
| T6 | 95.5 | 94.3 | 94.8 | 5.2 | 72.2 | 95.2 | 1.00 |
| T7 | 96.9 | 96.1 | 96.6 | 5.3 | 65.6 | 95.2 | 1.00 |
| T8 | 91.1 | 90.9 | 92.9 | 7.2 | 46.7 | 93.2 | 99.1 |
| T9 | 96.5 | 94.2 | 96.6 | 5.2 | 61.3 | 95.2 | 1.00 |
| Avg. | 95.3 | 94.6 | 95.4 | 5.4 | 70.1 | 95.0 | 99.4 |

TABLE XVIII: KVN+UM-PnP results on $bottle_0$ object

| bottle$_1$ | P2D | ADD(-S) | <2cm | MAE | 5c5d | AUC | AP70 |
|---|---|---|---|---|---|---|---|
| T0 | 97.1 | 98.8 | 98.8 | 4.5 | 94.3 | 96.0 | 1.00 |
| T1 | 91.8 | 97.1 | 97.6 | 7.2 | 79.0 | 93.3 | 1.00 |
| T2 | 93.9 | 96.7 | 96.7 | 5.5 | 90.7 | 95.0 | 1.00 |
| T3 | 79.6 | 94.6 | 95.3 | 8.3 | 76.6 | 92.1 | 1.00 |
| T4 | 84.3 | 92.9 | 93.3 | 8.9 | 78.0 | 91.5 | 99.4 |
| T5 | 96.5 | 98.1 | 98.1 | 5.0 | 94.1 | 95.5 | 1.00 |
| T6 | 92.8 | 97.4 | 97.7 | 5.6 | 90.5 | 94.8 | 1.00 |
| T7 | 94.3 | 96.0 | 96.0 | 6.2 | 86.4 | 94.3 | 1.00 |
| T8 | 89.8 | 97.2 | 97.2 | 6.6 | 83.7 | 93.9 | 1.00 |
| T9 | 89.1 | 97.2 | 97.2 | 6.4 | 81.7 | 94.1 | 1.00 |
| Avg. | 90.9 | 96.6 | 96.8 | 6.4 | 85.5 | 94.0 | 99.9 |

TABLE XIX: KVN+UM-PnP results on bottle$_1$ object

| mug$_0$ | P2D | ADD(-S) | <2cm | MAE | 5c5d | AUC | AP70 |
|---|---|---|---|---|---|---|---|
| T0 | 94.4 | 96.5 | 98.2 | 7.0 | 62.6 | 93.7 | 1.00 |
| T1 | 87.2 | 91.0 | 94.6 | 9.4 | 26.1 | 91.1 | 1.00 |
| T2 | 88.0 | 92.6 | 96.5 | 8.1 | 38.7 | 92.5 | 1.00 |
| T3 | 63.3 | 79.2 | 87.9 | 12.9 | 19.5 | 87.7 | 98.8 |
| T4 | 74.0 | 78.3 | 83.6 | 14.0 | 29.0 | 86.8 | 1.00 |
| T5 | 92.1 | 96.6 | 97.0 | 6.9 | 55.4 | 93.9 | 1.00 |
| T6 | 88.4 | 95.1 | 98.3 | 7.9 | 37.0 | 92.7 | 1.00 |
| T7 | 53.4 | 77.6 | 90.8 | 12.8 | 9.1 | 87.8 | 1.00 |
| T8 | 81.9 | 85.6 | 91.8 | 9.7 | 30.9 | 91.0 | 1.00 |
| T9 | 97.2 | 97.2 | 97.2 | 6.4 | 65.1 | 94.2 | 1.00 |
| Avg. | 82.0 | 89.0 | 93.6 | 9.5 | 37.3 | 91.1 | 99.9 |

TABLE XXIII: KVN+UM-PnP results on mug$_0$ object

| bottle$_2$ | P2D | ADD(-S) | <2cm | MAE | 5c5d | AUC | AP70 |
|---|---|---|---|---|---|---|---|
| T0 | 96.3 | 96.9 | 97.2 | 5.8 | 70.3 | 94.7 | 97.2 |
| T1 | 98.5 | 98.1 | 98.2 | 5.2 | 75.5 | 95.2 | 1.00 |
| T2 | 94.1 | 96.9 | 97.1 | 7.2 | 76.3 | 93.2 | 88.7 |
| T3 | 83.3 | 94.4 | 95.0 | 9.1 | 54.1 | 91.3 | 96.5 |
| T4 | 90.3 | 96.6 | 96.8 | 7.4 | 67.7 | 93.3 | 81.2 |
| T5 | 97.4 | 96.3 | 96.5 | 6.0 | 77.7 | 94.5 | 93.1 |
| T6 | 95.6 | 97.7 | 97.9 | 6.0 | 79.9 | 94.5 | 94.3 |
| T7 | 84.8 | 97.6 | 97.8 | 7.3 | 63.0 | 93.1 | 95.5 |
| T8 | 76.1 | 77.3 | 78.3 | 17.4 | 51.9 | 84.0 | 71.3 |
| T9 | 89.8 | 92.4 | 93.0 | 9.3 | 61.0 | 91.1 | 98.8 |
| Avg. | 90.6 | 94.4 | 94.8 | 8.1 | 67.7 | 92.5 | 91.7 |

TABLE XX: KVN+UM-PnP results on bottle$_2$ object

| mug$_1$ | P2D | ADD(-S) | <2cm | MAE | 5c5d | AUC | AP70 |
|---|---|---|---|---|---|---|---|
| T0 | 59.7 | 68.9 | 75.8 | 17.1 | 24.7 | 83.3 | 1.00 |
| T1 | 56.8 | 71.4 | 77.6 | 17.0 | 21.3 | 83.5 | 1.00 |
| T2 | 74.0 | 74.0 | 83.4 | 15.2 | 13.7 | 85.4 | 99.6 |
| T3 | 48.0 | 69.2 | 77.6 | 19.0 | 12.8 | 82.9 | 84.4 |
| T4 | 70.9 | 82.5 | 86.0 | 15.5 | 33.2 | 85.7 | 84.4 |
| T5 | 81.1 | 85.2 | 88.8 | 11.6 | 37.3 | 89.0 | 98.7 |
| T6 | 50.4 | 58.5 | 68.0 | 21.7 | 14.3 | 79.2 | 93.8 |
| T7 | 64.8 | 81.5 | 92.4 | 9.9 | 13.5 | 90.6 | 1.00 |
| T8 | 22.5 | 33.7 | 45.4 | 37.9 | 4.2 | 64.5 | 73.8 |
| T9 | 71.2 | 71.0 | 78.7 | 18.1 | 25.0 | 82.3 | 99.4 |
| Avg. | 59.9 | 69.6 | 77.4 | 18.3 | 20.0 | 82.6 | 93.4 |

TABLE XXIV: KVN+UM-PnP results on mug$_1$ object

| cup$_0$ | P2D | ADD(-S) | <2cm | MAE | 5c5d | AUC | AP70 |
|---|---|---|---|---|---|---|---|
| T0 | 99.2 | 98.9 | 99.2 | 5.1 | 55.7 | 95.4 | 1.00 |
| T1 | 99.4 | 99.3 | 99.4 | 4.3 | 64.5 | 96.3 | 1.00 |
| T2 | 99.4 | 98.7 | 99.1 | 4.9 | 46.0 | 95.6 | 1.00 |
| T3 | 97.0 | 91.9 | 97.5 | 8.2 | 28.2 | 92.2 | 82.1 |
| T4 | 98.3 | 96.8 | 98.1 | 6.0 | 57.9 | 94.6 | 97.5 |
| T5 | 99.6 | 99.4 | 99.6 | 4.7 | 54.4 | 95.7 | 1.00 |
| T6 | 94.6 | 96.4 | 98.0 | 6.6 | 24.6 | 93.9 | 93.3 |
| T7 | 91.0 | 91.4 | 94.3 | 8.7 | 41.3 | 92.1 | 95.3 |
| T8 | 91.1 | 87.9 | 91.9 | 8.9 | 17.7 | 91.6 | 88.4 |
| T9 | 98.6 | 97.6 | 98.7 | 5.8 | 31.0 | 95.0 | 1.00 |
| Avg. | 96.8 | 95.8 | 97.6 | 6.3 | 42.1 | 94.2 | 95.7 |

TABLE XXI: KVN+UM-PnP results on cup$_0$ object

| mug$_2$ | P2D | ADD(-S) | <2cm | MAE | 5c5d | AUC | AP70 |
|---|---|---|---|---|---|---|---|
| T0 | 97.6 | 97.2 | 97.5 | 6.4 | 72.4 | 94.6 | 1.00 |
| T1 | 97.3 | 95.8 | 97.1 | 8.7 | 65.9 | 92.1 | 1.00 |
| T2 | 81.4 | 86.2 | 89.9 | 12.4 | 45.1 | 88.6 | 1.00 |
| T3 | 88.7 | 90.8 | 93.3 | 10.4 | 56.8 | 90.8 | 99.7 |
| T4 | 80.5 | 88.2 | 91.6 | 10.8 | 22.9 | 89.9 | 97.8 |
| T5 | 96.3 | 95.9 | 97.4 | 7.7 | 57.2 | 93.2 | 1.00 |
| T6 | 91.1 | 89.9 | 93.2 | 10.8 | 32.1 | 90.2 | 1.00 |
| T7 | 87.1 | 85.9 | 90.1 | 10.6 | 41.7 | 90.2 | 1.00 |
| T8 | 90.5 | 93.1 | 94.6 | 9.9 | 37.0 | 91.4 | 96.7 |
| T9 | 92.7 | 95.8 | 97.4 | 7.9 | 54.2 | 93.0 | 1.00 |
| Avg. | 90.3 | 91.9 | 94.2 | 9.5 | 48.5 | 91.4 | 99.4 |

TABLE XXV: KVN+UM-PnP results on mug$_2$ object

| cup$_1$ | P2D | ADD(-S) | <2cm | MAE | 5c5d | AUC | AP70 |
|---|---|---|---|---|---|---|---|
| T0 | 98.2 | 98.2 | 98.2 | 5.0 | 72.3 | 95.5 | 1.00 |
| T1 | 98.8 | 98.8 | 98.8 | 4.1 | 91.1 | 96.4 | 1.00 |
| T2 | 99.2 | 99.2 | 99.2 | 3.5 | 77.2 | 97.0 | 1.00 |
| T3 | 85.2 | 95.8 | 96.4 | 8.0 | 64.4 | 92.4 | 99.1 |
| T4 | 87.4 | 96.3 | 96.7 | 7.5 | 47.4 | 92.9 | 99.2 |
| T5 | 98.8 | 98.8 | 98.8 | 4.2 | 88.3 | 96.3 | 1.00 |
| T6 | 99.4 | 99.4 | 99.4 | 3.8 | 75.6 | 96.6 | 1.00 |
| T7 | 97.3 | 99.0 | 99.0 | 4.4 | 71.1 | 96.1 | 1.00 |
| T8 | 86.6 | 91.1 | 92.1 | 11.6 | 45.3 | 88.8 | 98.1 |
| T9 | 95.3 | 94.0 | 94.2 | 8.0 | 75.8 | 92.4 | 1.00 |
| Avg. | 94.6 | 97.1 | 97.3 | 6.0 | 70.8 | 94.4 | 99.6 |

TABLE XXII: KVN+UM-PnP results on cup$_1$ object

| mug$_3$ | P2D | ADD(-S) | <2cm | MAE | 5c5d | AUC | AP70 |
|---|---|---|---|---|---|---|---|
| T0 | 92.7 | 97.4 | 98.0 | 7.3 | 70.3 | 94.0 | 98.4 |
| T1 | 97.1 | 97.2 | 97.3 | 7.5 | 87.0 | 94.1 | 98.1 |
| T2 | 72.6 | 84.5 | 86.2 | 13.2 | 59.0 | 88.4 | 93.4 |
| T3 | 95.7 | 96.3 | 96.5 | 7.1 | 83.1 | 94.0 | 1.00 |
| T4 | 75.0 | 86.9 | 89.0 | 12.6 | 34.2 | 88.8 | 50.0 |
| T5 | 93.3 | 96.9 | 97.2 | 10.2 | 78.8 | 92.8 | 97.5 |
| T6 | 96.2 | 98.7 | 98.8 | 5.4 | 89.0 | 95.4 | 99.7 |
| T7 | 90.0 | 92.1 | 92.9 | 9.4 | 67.7 | 91.7 | 1.00 |
| T8 | 75.9 | 81.6 | 82.9 | 15.3 | 48.7 | 85.5 | 73.8 |
| T9 | 88.7 | 93.0 | 93.7 | 12.9 | 59.2 | 90.6 | 95.3 |
| Avg. | 87.7 | 92.5 | 93.3 | 10.1 | 67.7 | 91.5 | 90.6 |

TABLE XXVI: KVN+UM-PnP results on mug$_3$ object

| $mug_4$ | P2D | ADD(-S) | <2cm | MAE | 5c5d | AUC | AP70 |
|---|---|---|---|---|---|---|---|
| T0 | 90.3 | 98.4 | 98.4 | 7.7 | 63.1 | 93.3 | 89.4 |
| T1 | 85.6 | 98.2 | 98.2 | 7.6 | 41.5 | 93.1 | 70.0 |
| T2 | 87.0 | 96.6 | 96.4 | 9.3 | 40.7 | 91.4 | 84.4 |
| T3 | 88.4 | 97.8 | 97.8 | 8.5 | 33.6 | 92.2 | 90.3 |
| T4 | 92.4 | 97.5 | 97.5 | 8.5 | 77.0 | 92.4 | 1.00 |
| T5 | 88.8 | 99.2 | 99.0 | 6.8 | 51.3 | 93.8 | 82.2 |
| T6 | 87.7 | 94.6 | 94.4 | 8.7 | 61.4 | 92.1 | 64.1 |
| T7 | 84.3 | 97.4 | 97.1 | 7.9 | 67.6 | 92.8 | 59.7 |
| T8 | 73.5 | 93.9 | 93.3 | 11.4 | 13.1 | 89.5 | 94.4 |
| T9 | 72.9 | 90.5 | 89.9 | 12.3 | 48.6 | 88.4 | 96.6 |
| Avg. | 85.1 | 96.4 | 96.2 | 8.9 | 49.8 | 91.9 | 83.1 |

TABLE XXVII: KVN+UM-PnP results on $mug_4$ object

| $mug_5$ | P2D | ADD(-S) | <2cm | MAE | 5c5d | AUC | AP70 |
|---|---|---|---|---|---|---|---|
| T0 | 93.2 | 96.1 | 96.1 | 8.7 | 88.7 | 92.7 | 92.2 |
| T1 | 91.8 | 91.6 | 91.6 | 13.7 | 83.5 | 88.3 | 1.00 |
| T2 | 87.3 | 91.4 | 91.4 | 14.2 | 71.6 | 87.8 | 75.8 |
| T3 | 86.0 | 91.7 | 91.4 | 14.3 | 72.3 | 87.5 | 99.4 |
| T4 | 93.1 | 95.5 | 95.5 | 10.1 | 75.2 | 91.3 | 53.0 |
| T5 | 91.0 | 92.7 | 92.7 | 12.3 | 83.8 | 89.7 | 79.1 |
| T6 | 91.8 | 93.3 | 93.2 | 11.5 | 61.1 | 90.5 | 65.0 |
| T7 | 92.3 | 95.1 | 95.0 | 12.2 | 71.6 | 90.1 | 75.0 |
| T8 | 87.2 | 92.5 | 92.3 | 11.7 | 65.8 | 89.9 | 90.6 |
| T9 | 91.9 | 93.2 | 93.1 | 11.5 | 85.7 | 89.9 | 84.4 |
| Avg. | 90.6 | 93.3 | 93.2 | 12.0 | 75.9 | 89.8 | 81.4 |

TABLE XXVIII: KVN+UM-PnP results on $mug_5$ object

| $mug_6$ | P2D | ADD(-S) | <2cm | MAE | 5c5d | AUC | AP70 |
|---|---|---|---|---|---|---|---|
| T0 | 87.8 | 93.9 | 93.6 | 12.3 | 61.4 | 89.6 | 99.6 |
| T1 | 94.9 | 96.7 | 96.1 | 7.7 | 88.2 | 93.2 | 1.00 |
| T2 | 92.6 | 93.4 | 92.8 | 11.9 | 71.1 | 89.3 | 1.00 |
| T3 | 77.6 | 84.8 | 83.9 | 18.7 | 57.5 | 83.2 | 98.1 |
| T4 | 77.1 | 86.1 | 84.5 | 16.8 | 39.3 | 83.8 | 1.00 |
| T5 | 88.8 | 93.1 | 93.0 | 12.8 | 63.7 | 90.2 | 1.00 |
| T6 | 89.5 | 92.2 | 92.0 | 11.9 | 63.1 | 89.4 | 1.00 |
| T7 | 86.8 | 88.6 | 87.6 | 13.3 | 66.7 | 87.7 | 1.00 |
| T8 | 81.9 | 90.9 | 89.5 | 14.8 | 40.3 | 87.2 | 95.0 |
| T9 | 88.5 | 94.0 | 93.8 | 11.8 | 36.2 | 89.9 | 1.00 |
| Avg. | 86.5 | 91.4 | 90.7 | 13.2 | 58.7 | 88.3 | 99.3 |

TABLE XXIX: KVN+UM-PnP results on $mug_6$ object

| $heart_0$ | P2D | ADD(-S) | <2cm | MAE | 5c5d | AUC | AP70 |
|---|---|---|---|---|---|---|---|
| T0 | 98.2 | 97.2 | 98.0 | 4.4 | 51.8 | 96.0 | 1.00 |
| T1 | 88.2 | 87.1 | 89.0 | 8.9 | 13.3 | 91.5 | 99.7 |
| T2 | 79.7 | 70.8 | 92.7 | 9.2 | 21.7 | 91.2 | 99.4 |
| T3 | 82.3 | 72.7 | 85.0 | 11.9 | 17.8 | 88.5 | 1.00 |
| T4 | 81.5 | 73.1 | 85.6 | 11.4 | 14.7 | 89.0 | 1.00 |
| T5 | 70.2 | 65.0 | 69.9 | 14.6 | 43.7 | 85.8 | 1.00 |
| T6 | 83.5 | 77.5 | 92.8 | 8.6 | 9.3 | 91.8 | 99.4 |
| T7 | 96.5 | 94.1 | 96.5 | 5.4 | 44.9 | 95.0 | 1.00 |
| T8 | 69.5 | 68.4 | 85.8 | 11.1 | 4.4 | 89.3 | 94.6 |
| T9 | 75.0 | 67.6 | 76.2 | 14.8 | 20.0 | 85.8 | 99.7 |
| Avg. | 82.5 | 77.3 | 87.2 | 10.0 | 24.2 | 90.4 | 99.3 |

TABLE XXX: KVN+UM-PnP results on $heart_0$ object

| $tree_0$ | P2D | ADD(-S) | <2cm | MAE | 5c5d | AUC | AP70 |
|---|---|---|---|---|---|---|---|
| T0 | 88.1 | 95.5 | 95.5 | 8.7 | 60.2 | 92.4 | 1.00 |
| T1 | 88.6 | 99.5 | 99.6 | 6.4 | 37.9 | 94.0 | 1.00 |
| T2 | 79.7 | 93.2 | 94.1 | 8.9 | 36.4 | 91.5 | 99.6 |
| T3 | 78.6 | 96.2 | 98.7 | 7.2 | 32.6 | 93.2 | 99.7 |
| T4 | 98.3 | 99.1 | 99.4 | 5.0 | 51.3 | 95.5 | 1.00 |
| T5 | 77.8 | 81.9 | 82.9 | 13.4 | 50.6 | 87.0 | 98.8 |
| T6 | 60.3 | 73.4 | 78.8 | 17.0 | 17.0 | 83.4 | 91.2 |
| T7 | 88.6 | 99.1 | 99.1 | 6.5 | 48.5 | 94.0 | 1.00 |
| T8 | 48.7 | 82.0 | 86.7 | 15.0 | 13.0 | 85.8 | 86.3 |
| T9 | 93.8 | 97.7 | 98.7 | 5.8 | 51.1 | 94.7 | 1.00 |
| Avg. | 80.3 | 91.8 | 93.4 | 9.4 | 39.8 | 91.1 | 97.6 |

TABLE XXXI: KVN+UM-PnP results on $tree_0$ object


\fi

\end{document}